\documentclass{article}

\usepackage{microtype}
\usepackage{graphicx}
\usepackage{subfigure}
\usepackage{booktabs} 

\usepackage{amsmath,amsfonts,bm}

\def\eqref#1{equation~\ref{#1}}

\def\1{\bm{1}}

\def\rs{{\textnormal{s}}}

\def\ry{{\textnormal{y}}}

\def\rvx{{\mathbf{x}}}

\def\vp{{\bm{p}}}

\def\vx{{\bm{x}}}

\def\vz{{\bm{z}}}

\def\mW{{\bm{W}}}

\DeclareMathAlphabet{\mathsfit}{\encodingdefault}{\sfdefault}{m}{sl}
\SetMathAlphabet{\mathsfit}{bold}{\encodingdefault}{\sfdefault}{bx}{n}

\def\sA{{\mathbb{A}}}

\DeclareMathOperator*{\argmax}{arg\,max}

\usepackage{bbm}
\usepackage{wrapfig}
\usepackage{xcolor}
\usepackage{amssymb}
\usepackage{amsthm}
\theoremstyle{lemma}

\usepackage{hyperref}

\usepackage[accepted]{icml2021}

\begin{document}

\twocolumn[
\icmltitle{Revisiting Explicit Regularization in Neural Networks for Well-Calibrated Predictive Uncertainty}



\begin{icmlauthorlist}
\icmlauthor{Taejong Joo}{est}
\icmlauthor{Uijung Chung}{est}
\end{icmlauthorlist}

\icmlaffiliation{est}{ESTsoft, Seoul, Republic of Korea}

\icmlcorrespondingauthor{Taejong Joo}{tjoo@estsoft.com}

\icmlkeywords{Machine Learning, ICML}

\vskip 0.3in
]



\printAffiliationsAndNotice{} 

\begin{abstract}
From the statistical learning perspective, complexity control via explicit regularization is a necessity for improving the generalization of over-parameterized models. However, the impressive generalization performance of neural networks with only implicit regularization may be at odds with this conventional wisdom. In this work, we revisit the importance of explicit regularization for obtaining well-calibrated predictive uncertainty. Specifically, we introduce a probabilistic measure of calibration performance, which is lower bounded by the log-likelihood. We then explore explicit regularization techniques for improving the log-likelihood on unseen samples, which provides well-calibrated predictive uncertainty. Our findings present a new direction to improve the predictive probability quality of deterministic neural networks, which can be an efficient and scalable alternative to Bayesian neural networks and ensemble methods.
\end{abstract}

\section{Introduction}
As deep learning models have become pervasive in real-world decision-making systems, the importance of calibrated predictive uncertainty is increasing. The calibrated performance refers to the ability to match its predictive probability of an event to the long-term frequency of the event occurrence \citep{dawid1982well}. Well-calibrated predictive uncertainty is of interest in the machine learning community because it benefits many downstream tasks such as anomaly detection \citep{malinin2019reverse}, classification with rejection \citep{lakshminarayanan2017simple}, and exploration in reinforcement learning \citep{gal2016dropout}. Unfortunately, neural networks are prone to provide poorly calibrated predictions. 

Bayesian neural networks have \textit{innate abilities} to represent the model uncertainty. Specifically, they express the probability distribution over parameters, in which uncertainty in the parameter space is determined by the posterior inference \citep{mackay1992practical,neal1993bayesian}. Then, we can quantify predictive uncertainty from aggregated predictions from different parameter configurations. From this perspective, \textit{deterministic} neural networks, which cannot provide such rich information, have been considered as lacking the uncertainty representation ability.



However, recent works \citep{lakshminarayanan2017simple,muller2019does,thulasidasan2019mixup} present a possibility of improving the quality of the predictive probability without changing the deterministic nature of neural networks. Specifically, they show that label smoothing \citep{szegedy2016rethinking}, mixup \citep{zhang2017mixup}, and adversarial training \citep{goodfellow2015explaining} significantly improve the calibration performance of neural networks. This direction is appealing because it can \textit{inherit} the scalability, computational efficiency, and surprising generalization performance of the deterministic neural networks, for which Bayesian neural networks often struggle \citep{wu2019deterministic,osawa2019practical,joo2020being}.


Motivated by these observations, we investigate a general direction from \textit{the regularization perspective} to mitigate the poorly calibrated prediction problem. Specifically, we introduce a probabilistic measure of calibration performance, which is lower bounded by the log-likelihood. Based on this theoretical insight, we explore several explicit regularization methods for improving the log-likelihood on the test dataset, which leads to well-calibrated predictive uncertainty. Our extensive empirical evaluation shows that explicit regularization significantly improves the calibration performance of deterministic neural networks. 

Our contributions can be summarized as follows:
1) our theoretical analysis clearly connects the log-likelihood with calibration performance, which provides a direction to improve the quality of predictive uncertainty. 
2) we present a new direction for providing well-calibrated predictive uncertainty, which is readily applicable and computationally efficient;
3) our findings provide a novel view on the role and importance of explicit regularization in deep learning.

\section{Background}
We consider a classification problem with i.i.d. training samples $\mathcal{D} = \left\lbrace (\vx^{(i)}, y^{(i)})  \right\rbrace_{i=1}^N$ drawn from unknown distributions $P_{\rvx, \ry}$ whose corresponding tuple of random variables is $(\rvx,\ry)$. We denote $\mathcal{X}$ as an input space and $\mathcal{Y}$ as a set of class labels $\{1,2, \cdots, K \}$.

We define a classifier by a composition of a neural network and a softmax function; specifically, let $f^{\mW} : \mathcal{X} \rightarrow \mathcal{Z}$ be a neural network with parameters $\mW$ where $\mathcal{Z} = \mathbb{R}^K$ is a logit space. On top of the logit space, the softmax $\sigma: \mathbb{R}^K \rightarrow \triangle^{K-1}$ normalizes the exponential of logits:
\begin{equation}
    \phi^{\mW}_k(\vx) = \frac{\exp(f^{\mW}_k(\vx))}{ \sum_i \exp(f^{\mW}_i(\vx))}
\end{equation}
where we let $\phi^{\mW}_k(\vx) = \sigma_k (f^{\mW}(\vx))$ for brevity.

We often interpret $\phi^{\mW}(\vx)$ as the predictive probability that the label of $\vx$ belongs to class $k$ \citep{bridle1990probabilistic}.  Then, we train a neural network by maximizing the log-likelihood with stochastic gradient descent (SGD) \citep{robbins1951stochastic} or its variants. Here, we note a standard result that the maximum of the log-likelihood is achieved if and only if $\phi^{\mW}(\rvx)$ perfectly matches $p(\ry|\rvx)$.


\subsection{Calibrated Predictive Uncertainty}
A model with well-calibrated predictive uncertainty provides trustworthy predictive probability whose confidence\footnote{Throughout this paper, the confidence at $\vx$ refers to $\max_k \phi_{k}^{\mW}(\vx)$, which is different from the confidence in the statistics literature.} aligns well with its expected accuracy.
Formally, a well-calibrated model satisfies the following condition: 
$\forall \vp \in \triangle^{K-1}, k \in \{1, 2, \cdots, K \},$
\begin{equation} \label{eq:calibration}
    p(\ry = k |\phi^{\mW}(\rvx) = \vp ) = \vp_k.
\end{equation}
Here note that the calibrated model does not necessarily be ones achieving zero log-likelihood.



Given a finite dataset $\mathcal{D}$, we cannot evaluate the calibration condition in \eqref{eq:calibration}. Therefore, in practice, expected calibration error (ECE) \citep{naeini2015obtaining} is widely used as a calibration performance measure. ECE on $\mathcal{D}$ can be computed by binning predictions into $M$ groups based on their confidences and then averaging their calibration scores by:
\begin{equation}
    \sum_{i=1}^{M} \frac{| \mathcal{G}_i|}{|\mathcal{D}|}| \text{acc}(\mathcal{G}_i) - \text{conf}(\mathcal{G}_i)|
\end{equation} 
where $\mathcal{G}_i = \left\lbrace \vx : \frac{i}{M} < \max_k \phi_{k}^{\mW}( \vx) \leq \frac{i+1}{M}, \vx \in \mathcal{D} \right\rbrace$; $\text{acc}(\mathcal{G}_i)$ and $\text{conf}(\mathcal{G}_i)$ are average accuracy and confidence of predictions in group $\mathcal{G}_i$, respectively.

Another common way to evaluate the predictive uncertainty's quality is based on predictive entropy, which assesses how well the model is aware of its ignorance. For example, predictive entropy is measured on misclassified or out-of-distribution (OOD) samples; that is, compute $- \sum_k \phi_k^{\mW}(\vx^\prime) \log \phi_k^{\mW}(\vx^\prime)$ for some ignorant sample $\vx^\prime$. For a well-calibrated model, we expect the high predictive uncertainty on such samples, i.e., the answer ``I don't know.''

\begin{figure}
\begin{subfigure}
  \centering
  \includegraphics[width=.48\linewidth]{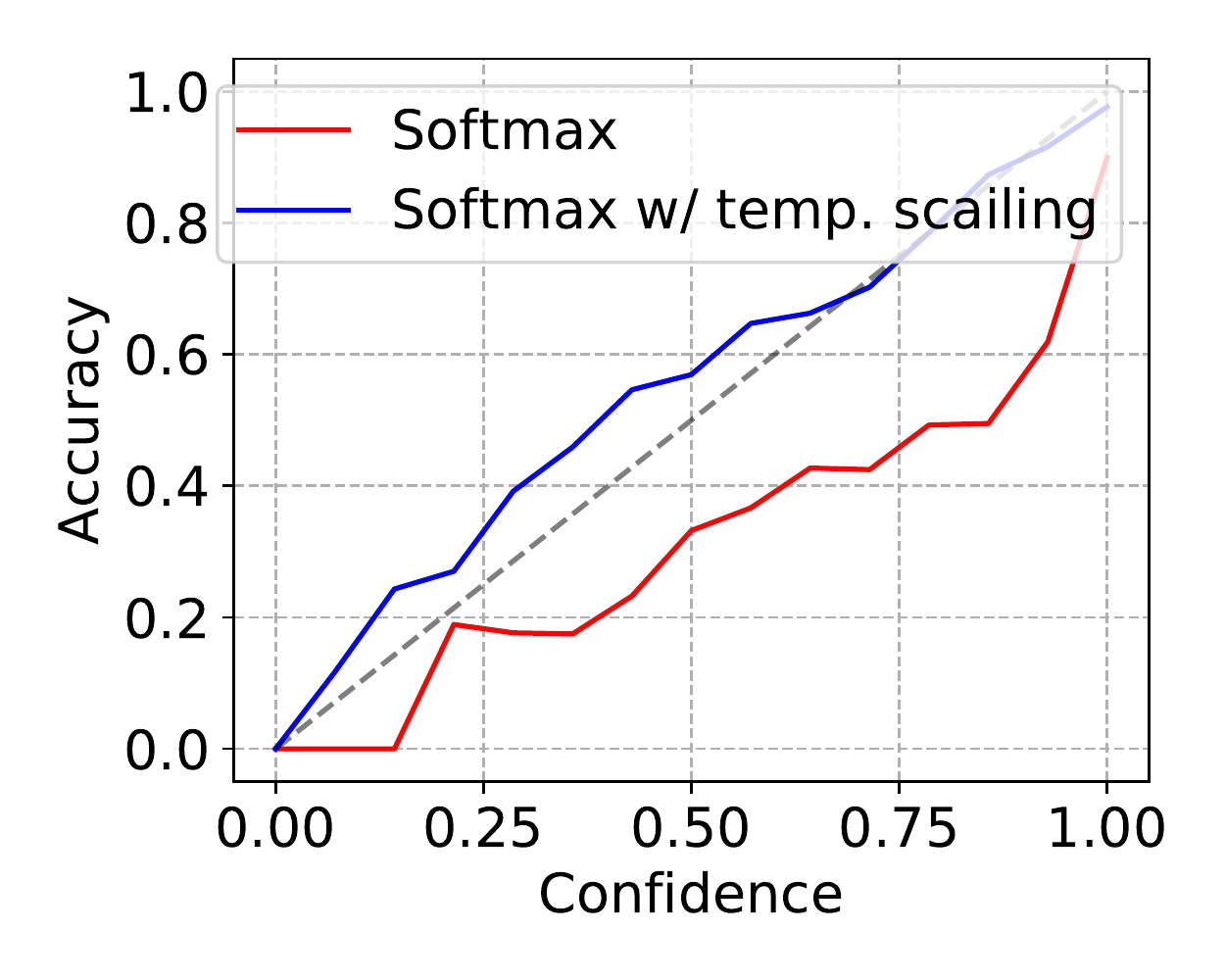}
\end{subfigure}
\hfill
\begin{subfigure}
  \centering
  \includegraphics[width=.48\linewidth]{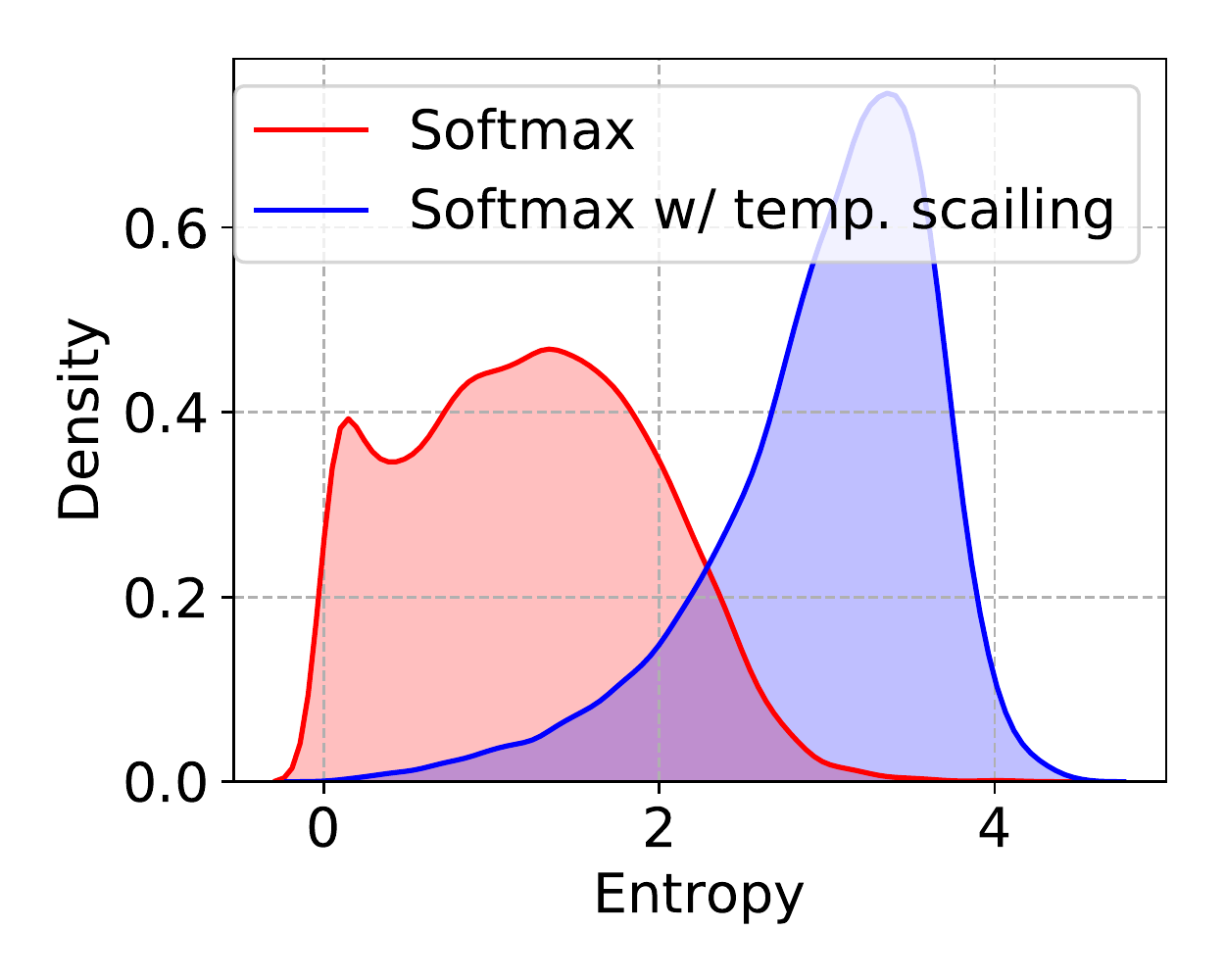}
\end{subfigure}
\vspace{-10pt}
\caption{Reliability curve (left) and predictive uncertainty on out-of-distribution samples (right) of ResNet.}
\label{motivate_example}
\end{figure}

Unfortunately, it is widely known that neural networks produce uncalibrated predictive uncertainty. For example, Figure~\ref{motivate_example} illustrates the uncalibrated predictive behavior of ResNet \citep{he2016identity}: its confidence tends to higher than its accuracy (Figure~\ref{motivate_example} (left)); also, it provides low predictive entropy on OOD samples (Figure~\ref{motivate_example} (right)). Due to this problem, interpretation of $\phi^{\mW}(\rvx)$ as the ``predictive probability'' is often considered as implausible \citep{gal2016dropout}.

\subsection{Temperature Scaling}
One simple yet effective solution for improving the predictive probability's quality is to use post-hoc calibration methods. Among them, temperature scaling \citep{guo2017calibration} adjusts the smoothness of the softmax outputs for maximizing the log-likelihood on a holdout dataset $\mathcal{D}^\prime$:
\begin{equation} \label{eq:temp_scale}
    \max_{\tau} \sum_{(\vx, y) \in \mathcal{D}^\prime} \log\frac{\exp(f_{y}^{\mW}(\vx) / \tau)}{\sum_j \exp(f_{j}^{\mW}(\vx) / \tau)}
\end{equation}
where $\mW$ is a fixed pretrained weight and $\tau$ is a temperature controlling the smoothness of the softmax output. 

Figure~\ref{motivate_example} illustrates the effectiveness of the temperature scaling on providing a better calibrated predictive uncertainty: the confidence closely matches its actual accuracy and the predictive entropy on OOD samples significantly increases. However, the post-hoc calibration methods have some inherent disadvantages: the performance is sensitive to the quality of the holdout dataset; also, they cannot utilize full training data, which could result in lower generalization performance. For these reasons, we explore an alternative approach that does not require the post-hoc calibration. 

\begin{figure*}
\centering
\begin{subfigure}[]
  {\includegraphics[width=.23\linewidth]{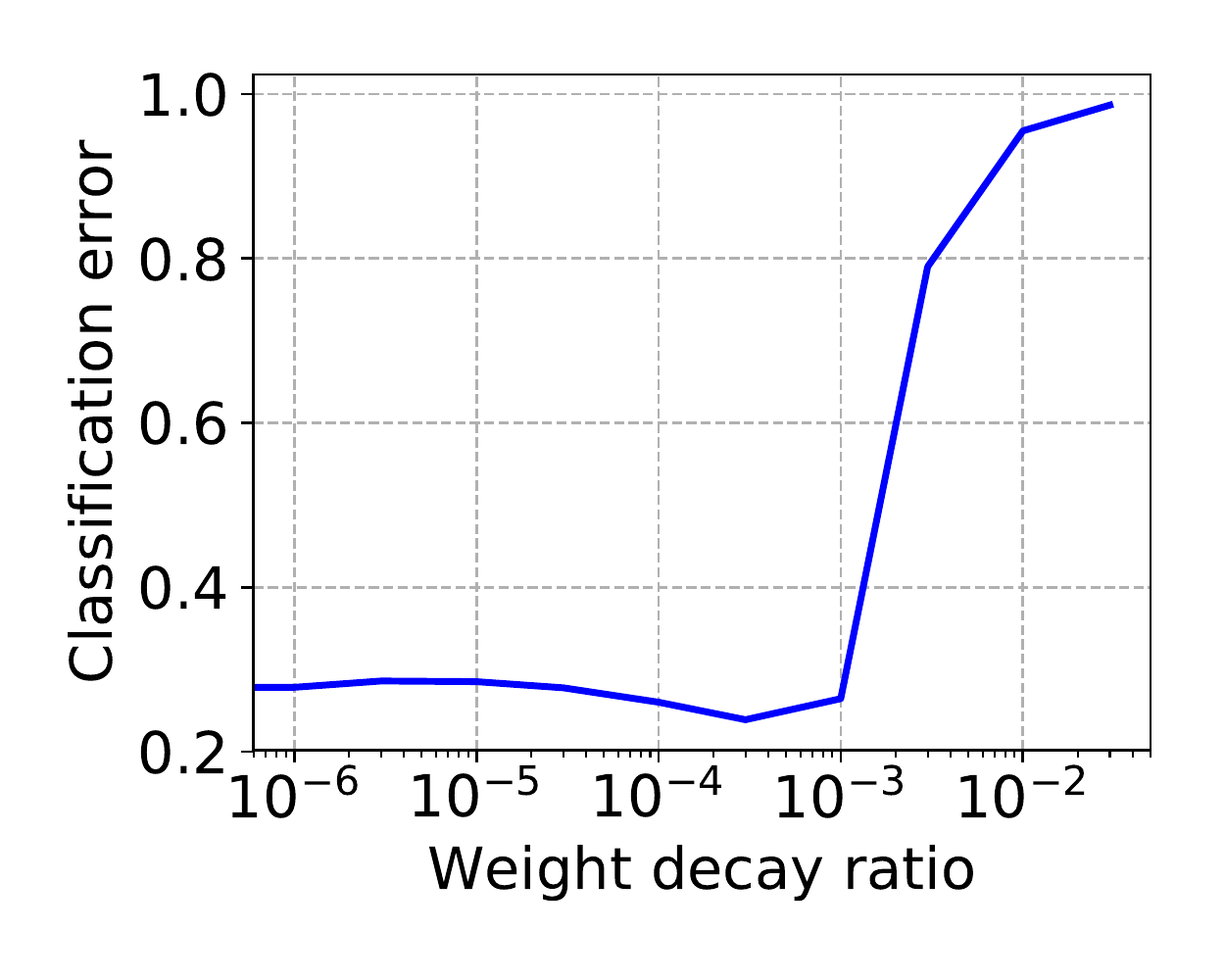}}
\end{subfigure}
\begin{subfigure}[]
  {\includegraphics[width=.23\linewidth]{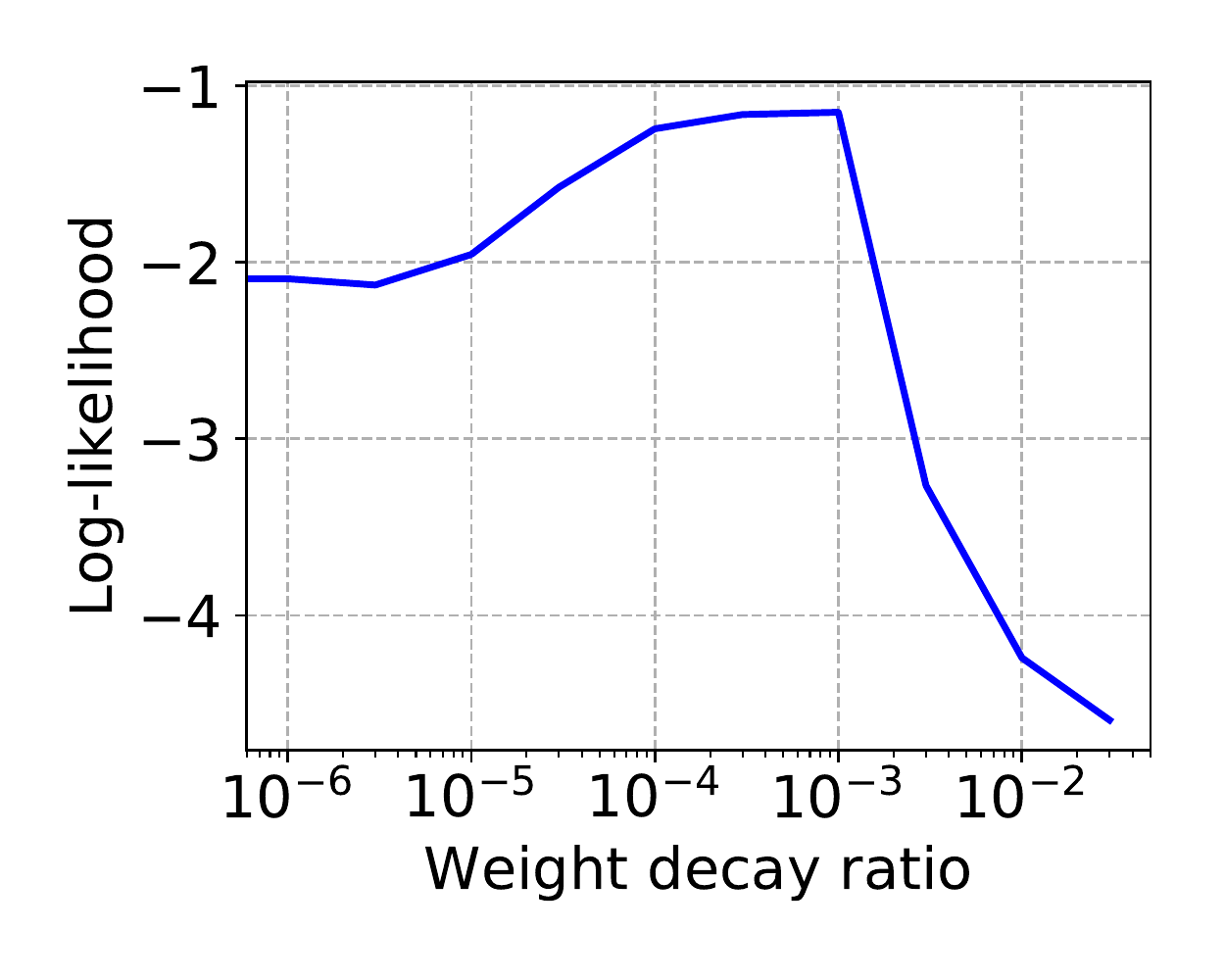}}
\end{subfigure}
\begin{subfigure}[]
  {\includegraphics[width=.27\linewidth]{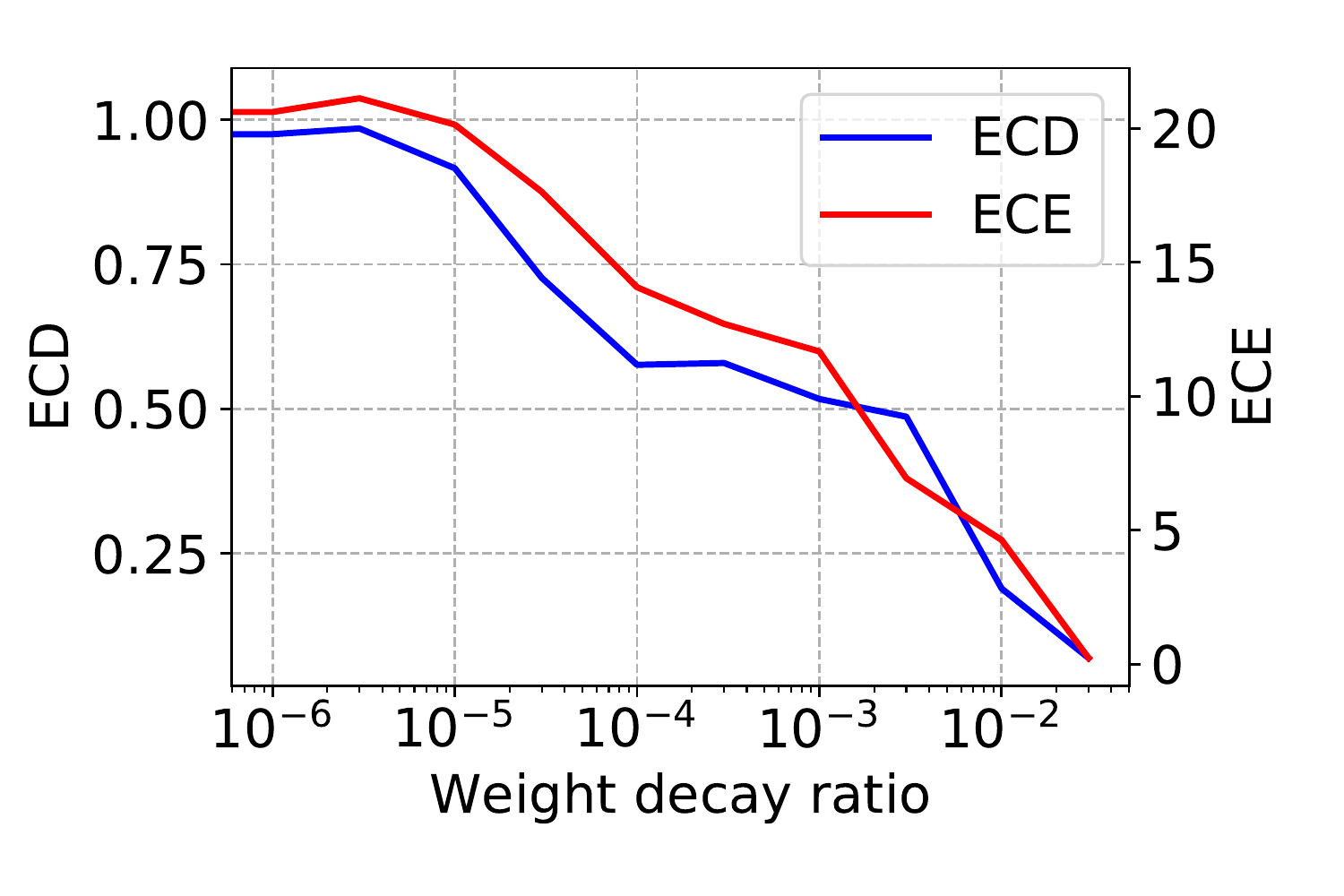}}
\end{subfigure}
\begin{subfigure}[]
  {\includegraphics[width=.22\linewidth]{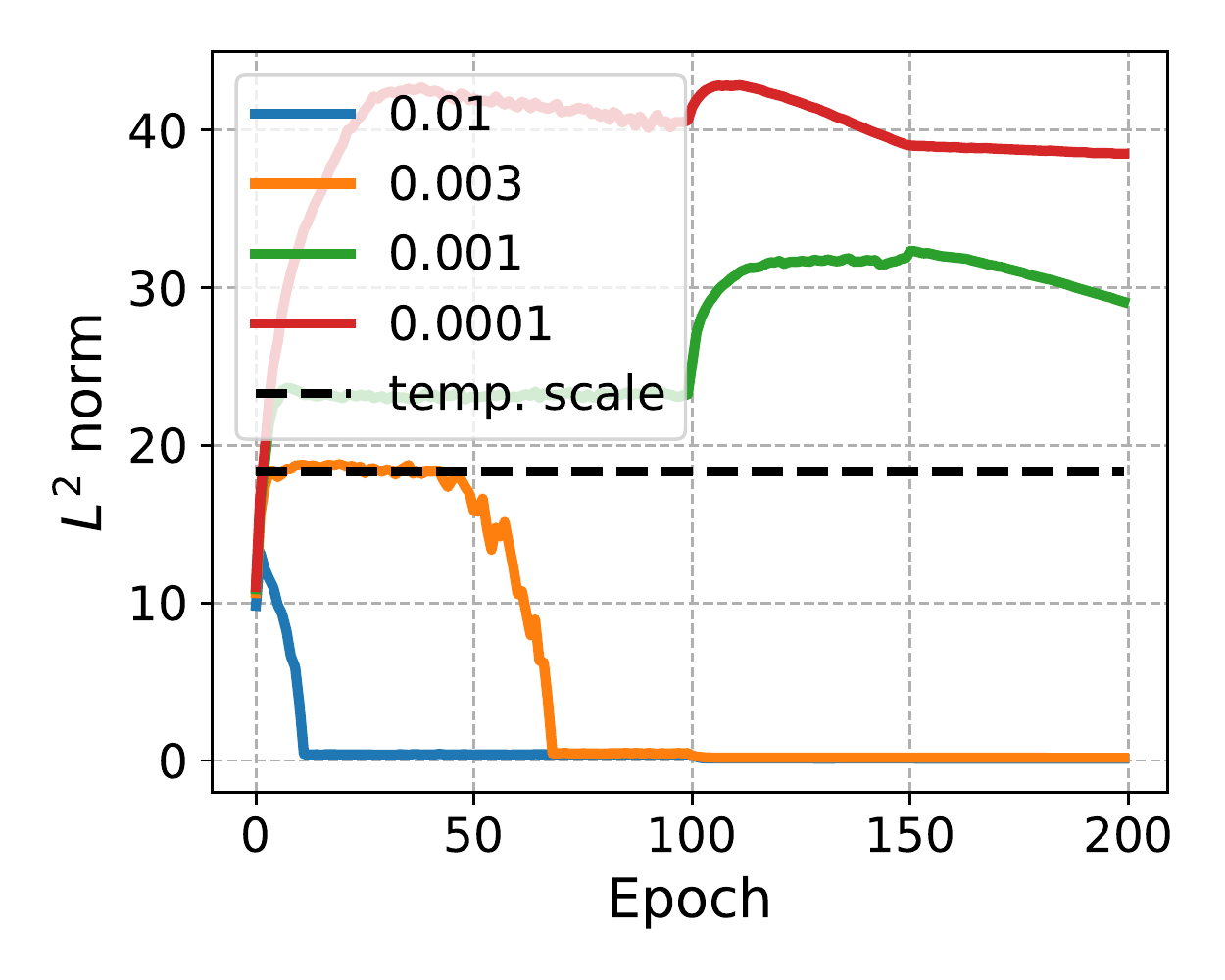}}
\end{subfigure}
\vskip -0.2in
\caption{Impacts of the weight decay rate on accuracy (a), log-likelihood (b), calibration performance (c), and and the $L^2$ norm (d). }
\label{init_wd_test}
\end{figure*}

\section{Motivation} \label{sec:motivation}
Motivated by the temperature scaling, we closely analyze how the log-likelihood is related to the calibration performance. To this end, let $\rs_{\rvx}$ be a binary random variable indicating whether a model correctly classifies a sample $\rvx$; $ p(s_\rvx) = p(\ry = \argmax_k \phi^{\mW}_{k}(\rvx))$ where $\ry$ is a true label of $\rvx$. Then, we can derive the following inequality by using the law of total expectation: 
\begin{equation} \label{eq:decompose_ll}
\begin{split}
    \mathbb{E}_{\rvx,\ry}[\log \phi^{\mW}_{\ry}(\rvx)]  
    = \mathbb{E}_{\rvx} \left[
    \mathbb{E}_{\rs} \left[
    \mathbb{E}_{\ry} [
                \log \phi^{\mW}_{\ry}(\rvx) | \rs_{\rvx}, \rvx
    ]|\rvx \right] \right]
    \\
    \leq \mathbb{E}_{\rvx}[p(s_{\rvx})  \log \phi^{\mW}_{m_{\rvx}}(\rvx)  +  p(\neg s_{\rvx}) \log (1 - \phi^{\mW}_{m_{\rvx}}(\rvx) )]
    \end{split}
\end{equation}
where  $m_{\rvx}= \argmax_k \phi^{\mW}_k(\rvx)$ and the inequality comes from the fact that $\mathbb{E}_{\ry} \left[\log \phi^{\mW}_{\ry}(\rvx) | \neg s_\rvx, \rvx \right] \leq \max_{k \neq m_{\rvx}} \log \phi^{\mW}_{k}(\rvx) \leq \log \left(1 - \phi^{\mW}_{m_{\rvx}}(\rvx) \right) $.

    

Consider we apply the inequality in \eqref{eq:decompose_ll} to the log-likelihood on test samples $\mathcal{D}^T$. Suppose we group predictions based on the confidence such that $\vx, \vx^\prime \in \mathcal{G}_k^\epsilon \rightarrow | \log \phi^\mW_{m_\vx} - \log \phi^\mW_{m_{\vx^\prime}} | < \epsilon $ for arbitrary small $\epsilon$ and some $k$. Then, we can approximate the upper bound by:
\begin{multline}
\hat{\mathbb{E}}_{\mathcal{D}^T}\left[p(s_{\rvx})  \log \phi^{\mW}_{m_{\rvx}}(\rvx)  +  p(\neg s_{\rvx}) \log (1 - \phi^{\mW}_{m_{\rvx}}(\rvx) )\right] \\
    \approx - \sum_{i = 1}^{M^\epsilon} \frac{|\mathcal{G}^\epsilon_i|}{N^T} CE( \text{acc}(\mathcal{G}^\epsilon_i) \parallel \text{conf}(\mathcal{G}^\epsilon_i)) 
\end{multline}
where $\hat{\mathbb{E}}_{\mathcal{D^T}}[\cdot ]$ is the empirical mean on $\mathcal{D}^T$, $|\mathcal{G}^\epsilon_i|$ is the number of samples in $\mathcal{G}^\epsilon_i$, $N^T$ is the number of test samples,  $M^\epsilon$ is a number of groups given $\epsilon$, and $CE(\cdot \parallel \cdot)$ is a cross-entropy.

This upper bound can be thought of as the probabilistic measure of calibration performance, and we refer to its negative value as expected calibration divergence (ECD). Specifically, consider a neural network that produces an answer with probability $\text{conf}(\mathcal{G}^\epsilon_i)$ and refuses to answer otherwise. Then, ECD becomes the expected divergence between the model's ability to correctly predict the sample and the model's willingness to answer. 






The above analysis suggests that improving the log-likelihood on test samples can lead to a better-calibrated model. This means that we can focus on \textit{generalization of the training objective} for improving the calibration performance. To achieve this, we explore explicit regularization techniques for improving the log-likelihood on unseen samples. We remark that our approach is fundamentally different from previous regularization research in deep learning concentrating on the effects of regularization on generalization performance, which treats the log-likelihood as a surrogate loss for improving accuracy.

\section{Regularization by Strong Weight Decay}
\label{sec_wd}
Weight decay \citep{krogh1992simple} is one of the most widely used explicit regularization techniques, which is already involved in many benchmark models. However, we conjecture that the decay rate in benchmark training strategies, e.g., 0.0001 in ResNet, is too small to induce a regularization effect\footnote{Here, we apply the weight decay in the form of \citet{loshchilov2017decoupled} that shrink weights by a fixed-rate at each training step to ensure proper regularization effects on weights under adaptive gradient-based optimizers.}. Therefore, we investigate the impacts of a large weight decay rate on log-likelihood and calibration performance.

Figure~\ref{init_wd_test} illustrates the impact of the weight decay rate $\lambda$ on generalization performance (a), log-likelihood (b), and calibration performance (c). Specifically, in the region of $ \lambda \leq 0.001$, a stronger weight decay improves the log-likelihood. Also, as in the theoretical analysis in Section~\ref{sec:motivation}, the log-likelihood improvement results in better calibration performance in terms of both ECD and ECE. However, these improvements become \textit{inversely} proportional to the accuracy improvement when $\lambda > 0.001$ (which is only 10x larger than a base decay ratio). 


To more thoroughly investigate this undesirability, we further analyze the impact of weight decay on the $L^2$ function norm of $f^\mW$. Here, the $L^2$ norm of $f^\mW$ is defined as $\parallel f^{\mW} \parallel_2 = \left( \int |f^{\mW}(\vx)|^2 dP_{\rvx}(\vx) \right)^{1/2}$, which we will discuss with more detail in Section \ref{sec:method}. In this work, we compute $L^2$ norm by the Monte-Carlo approximation since we do not have access to a data-generating distribution $P_{\rvx}(\vx)$.

To set a desired value of $L^2$ norm, we first note that temperature scaling modifies only $L^2$ norm without affecting the accuracy (cf. \eqref{eq:temp_scale}). We also note that moderate changes in $\lambda$ do not greatly impact the generalization error (cf. Figure~\ref{init_wd_test} (a)). Therefore, the calibrated $L^2$ norm obtained by applying temperature scaling on the test set $\mathcal{D}^{T}$ can be thought of as the target complexity for maximizing the log-likelihood given certain generalization performance of $f^{\mW}$, for which we want to achieve by varying $\lambda$.


In Figure~\ref{init_wd_test} (d), we monitor $ \parallel f^{\mW} \parallel_2 $ under all considered weight decay rates, which shows that the SGD with various decay rates finds only \textit{a trivial solution} or \textit{an infeasible solution} to the following optimization problem:
\begin{gather} \label{eq:opt_prob}
    \max_{\mW} \hat{\mathbb{E}}_{\mathcal{D}} \left[
        \log \phi_{\ry}^{\mW}(\rvx)
    \right]   
    \quad \nonumber 
    \\ 
    \text{s.t.} \parallel f^{\mW} \parallel_2 \leq \parallel f^{{\mW^\prime}} / \tau^* \parallel_2
\end{gather}
where $\mW^\prime$ is the trained weights with $\lambda = 0.0001$, and $\tau^{*} = \argmax_{\tau} \hat{\mathbb{E}}_{\mathcal{D}^{T}} \left[ \log\left\lbrace \sigma ( f^{{\mW^\prime}}_{\ry}(\rvx) /\tau )\right\rbrace \right]$. 
Specifically, Figure~\ref{init_wd_test} (d) shows that the $L^2$ norm goes to zero under the decay rate $\lambda \geq  0.003$, which means that all weights collapse to zero, i.e., the trivial solution. This happens when the magnitude of the decay rate overwhelms the magnitude of the weight update by the gradient of the log-likelihood, e.g., at approximately the epoch 50 under $\lambda=0.003$ (Figure~\ref{init_wd_test} (d)). In the case of $\lambda \leq 0.001$, SGD with $\lambda$ does not suffer from the weight collapse, but the scale of the $L^2$ norm under such rates is significantly higher than $\parallel f^{{\mW^\prime}} / \tau^{*} \parallel_2$, which corresponds to infeasible solutions (Figure~\ref{init_wd_test} (d)). 

We may overcome this undesirability by using a dynamic weight decay rate for each parameter that adjusts the decay rate corresponding to its overall gradient scale, which can prevent the weight collapse while reducing the $L^2$ norm. However, this approach requires a rule for the decay rate update (or a meta-learning algorithm with additional parameters). Thus, we will explore alternative forms of explicit regularization that directly penalize the function complexity.

\section{Direct Regularization of Function Complexity} \label{sec:method}

\subsection{Regularization in the Function Space}
The first approach regards $f^{\mW}$ as an element of $L^p(\mathcal{X})$ space. $L^p$ space is the space of measurable functions with the norm:
\begin{equation} \label{eq:lp_def}
    \parallel f^{\mW} \parallel_p = \left(\int_{\mathcal{X}} | f^{\mW} (\vx) |^p dP_{\rvx}(\vx) \right)^{1/p} < \infty.
\end{equation}
As aforementioned, we use Monte-Carlo approximation for computing $L^p$ norm, which can be computed as $\parallel f^{\mW} \parallel_p^p \approx \frac{1}{m} \sum_{i,j} | f^{\mW}_j (\vx^{(i)}) |^p$. In this paper, we examine $ \parallel f^{\mW} \parallel_1 $ and $ \parallel f^{\mW} \parallel_2^2 $ regularization losses.

\begin{figure}
  \centering
    \hfill
    \begin{subfigure}
     {\includegraphics[width=0.23\textwidth]{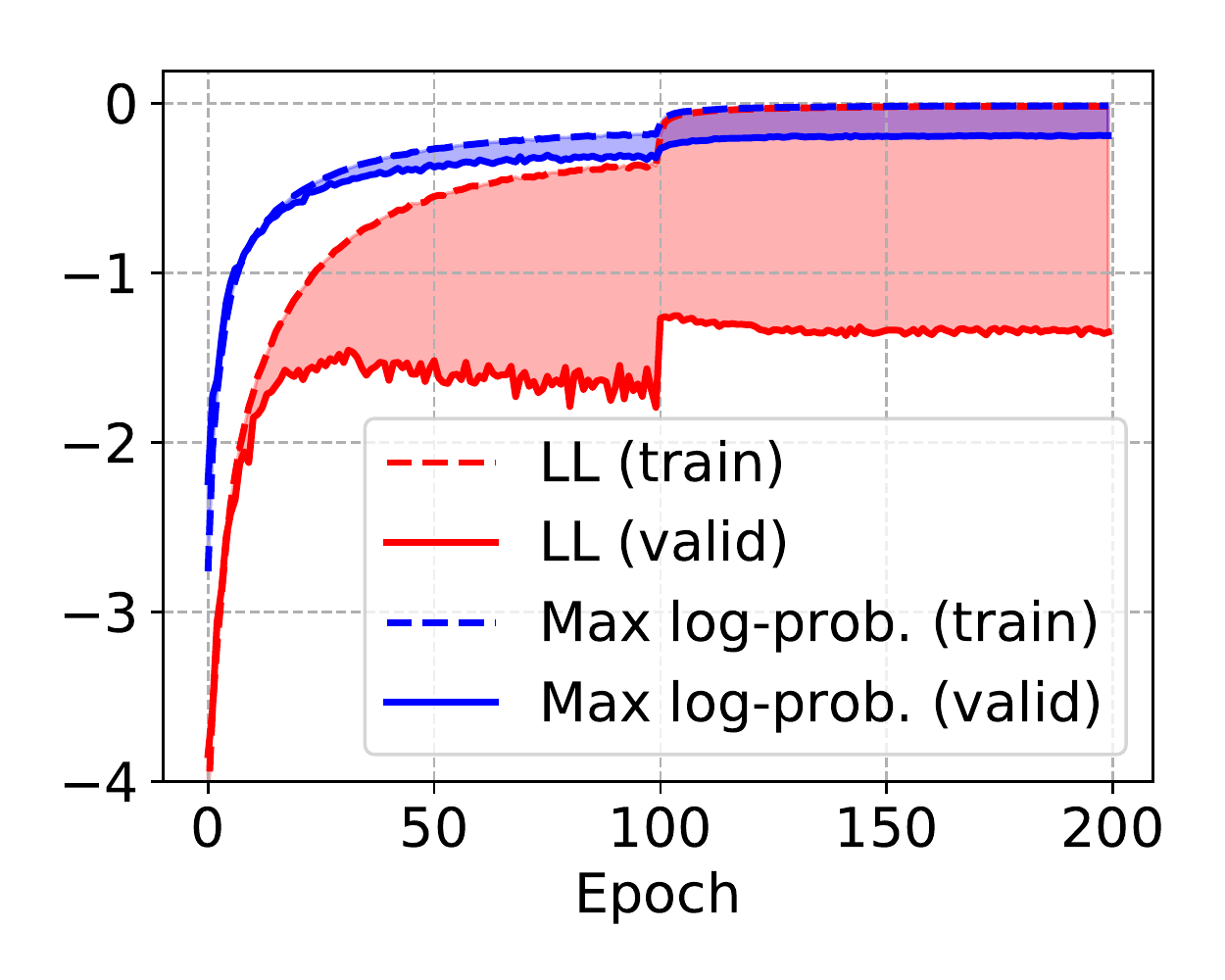}}
    \end{subfigure}
    \hfill
    \begin{subfigure}
     {\includegraphics[width=0.23\textwidth]{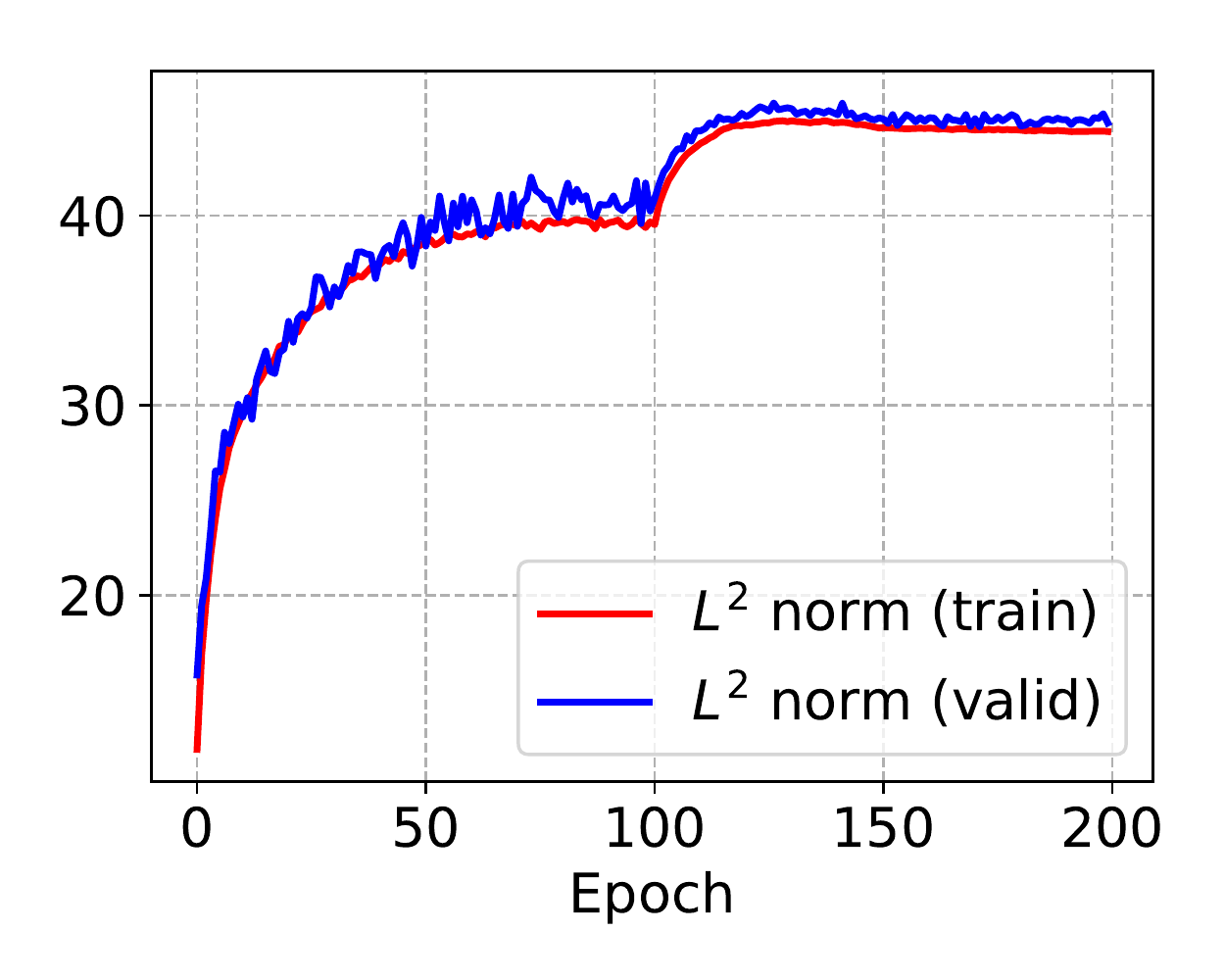}}
    \end{subfigure}
    \vskip -0.1in
    \caption{Monitoring changes in the behavior of ResNet during training on CIFAR-100.
    LL indicates the log-likelihood $\mathbb{E}_{\rvx, \ry}[\log \phi^{\mW}_{\ry}(\rvx)] $ and max log-prob indicates $\mathbb{E}_{\rvx}[\log \phi^{\mW}_{m_{\rvx}}(\rvx)] $. 
    }
    \label{train_analysis}
\end{figure}

We note that the norm is computed with respect to the input generating distribution $P_{\rvx}$. Therefore, if the function complexity approximated with the training distribution does not reflect its true complexity on $P_{\rvx}$, regularizing the $L^p$ norm during training could be meaningless. Fortunately, we observe that function's properties, which depends only on $\rvx$, evaluated on two different datasets drawn from the identical distribution are very close to each other compared to values that depend on the external randomness $\ry | \rvx$.
For example, the empirical mean of the maximum log-probability $\log \phi^{\mW}_{m_{\rvx}}(\rvx)$ (Figure~\ref{train_analysis} (left)) and $L^2$ norm of $f^{\mW}$ (Figure~\ref{train_analysis} (right)) on training samples are significantly similar to those values on unseen samples compared to the log-likelihood (Figure~\ref{train_analysis} (left)).

\begin{table*}
\caption{Experimental results under various regularization methods on ResNet. Arrows on the metrics represent the desirable direction. $\lambda^*$ represents the best hyperparameter.
}
\label{benchmark_result}
\begin{center}
\begin{small}
\begin{tabular}{llllll|lllll}
\toprule
& \multicolumn{5}{c |}{CIFAR-10} & \multicolumn{5}{c}{CIFAR-100} \\
method & Acc $\uparrow$ & NLL $\downarrow$ & ECE $\downarrow$ & $\parallel f^{\mW} \parallel_2$  & $\lambda^*$ & Acc $\uparrow$ & NLL $\downarrow$ & ECE $\downarrow$ & $\parallel f^{\mW} \parallel_2$ & $\lambda^*$
\\
\midrule
ResNet-50 &   94.17    &   0.29 &   4.06 &  20.12 & -  & 74.64  &   1.31 & 13.95 &   43.81   &  -  \\
+ $\parallel f^{\mW} \parallel_1$ &   94.32   &   0.25 &     2.89    &     6.81  &  0.01  &   76.28 &   1.27 &   7.77 &   9.07  &  0.01  \\
+ $\parallel f^{\mW} \parallel_2^2$ &   94.38   &   0.23  &   3.27  &     8.19  &  0.003 &   75.84 &   1.07 &  5.52  &  10.55   &  0.01  \\
+ $SW_1(\mu^{\mW}_{\mathcal{D}^\prime}, \nu)$ &   94.46  &   0.23 &   2.12  &  5.4  &  0.001 &   76.27 &   1.1  &   7.02   &  9.67 &  0.01   \\
+ PER &   94.30 &   0.23 &   2.89   &  6.94  &  0.03 & 76.23  &   1.15 &   4.67 &   7.56 &  1.0  \\
\midrule
VGG-16 &   92.97    &   0.35 &   4.96 &  9.62 & - &   71.96 &   1.4 &   16.9 &    22.98  &  -  \\
+ $\parallel f^{\mW} \parallel_1$ &   93.07  &   0.33  &   4.1  &    6.62  &  0.01 &  72.71 &   1.44 &     11.37  &  9.27   &  0.03  \\
+ $\parallel f^{\mW} \parallel_2^2$ &   93.06 &   0.31 & 4.58 &    7.44   &  0.003 &   72.68 &   1.31 &    10.79 &    9.28    &  0.01  \\
+ $SW_1(\mu^{\mW}_{\mathcal{D}^\prime}, \nu)$ &   93.13 &   0.29  &    1.9 &     5.4  &  0.001 & 72.26  &   1.35  &     12.59  &   9.72  &  0.03 \\
+ PER &    93.1  &   0.31 &   4.79  &     8.49  &  0.003  &  72.89 &   1.34 &   6.47 &   7.37 &  1.0 \\
\bottomrule
\end{tabular}
\end{small}
\end{center}
\vskip -0.1in
\end{table*}

\subsection{Regularization in the Probability Distribution Space}
The second approach regards $f^{\mW} (\rvx)$ as a random variable and then minimizes its distance to the bell-shaped target distribution (e.g., the standard Gaussian). Since the bell-shaped distribution has a peak mass at zero, this regularization encourages logits to have small values, making $\parallel f^{\mW} \parallel_2$ small. It also has additional regularization effects such as decorrelating logits and producing high predictive entropy.



As a metric in the probability distribution space, we use the sliced Wasserstein distance of order one because of its computational efficiency and ability to measure the distance between probability distributions with different supports, which is useful when dealing with the empirical distribution. We refer to \citet{peyre2019computational} for more detailed explanations about this metric.

Given minibatch samples $\mathcal{D}^{\prime} = \{ \vx^{(i)}\}^{m}_{i=1} $, we denote an empirical measure of logits as $\mu_{\mathcal{D}^{\prime}}(\sA) := \frac{1}{m} \sum_i \mathbbm{1}_{\sA} (f^{\mW}(\vx^{(i)})) $ and the standard Gaussian measure on $\mathcal{Z}$ as $\nu^{\mathcal{Z}} (\sA) := \frac{1}{2\pi^{K/2}} \int_{\sA} \exp \left( -\frac{1}{2} \parallel \vz \parallel^2 \right) d\vz $. 
Then, the sliced Wasserstein distance between them is given by:
\begin{equation} \label{eq:sw_compute}
    SW_1(\mu_{\mathcal{D}^{\prime}}, \nu^{\mathcal{Z}}) 
    = \int_{\mathbb{S}^{K-1}} 
        \int_{\mathbb{R}} 
        \left| F_{\mu_{\mathcal{D}^\prime}^{\theta}}(x) - F_{\nu^\mathbb{R}}(x) \right|  dx d \lambda(\theta)
\end{equation}
where  $\vz^{(i)} = f^{\mW}(\vx^{(i)})$, $\mu_{\mathcal{D}^\prime}^{\theta}$ is a measure obtained by projecting $\mu_{\mathcal{D}^{\prime}} $ at angle $\theta$, $F_{\mu}$ is a cumulative distribution function of a probability measure $\mu$, and $\lambda$ is a uniform measure on the unit sphere $\mathbb{S}^{K-1}$. 
In this work, the integration over $\mathbb{S}^{K-1}$ is evaluated by Monte-Carlo approximation with 256 number of evaluations, following \citet{joo2020regularizing}.

We also consider projected error function regularization (PER) \citep{joo2020regularizing} that simplifies the computation of $SW_1(\mu_{\mathcal{D}^{\prime}}, \nu^{\mathcal{Z}}) $ by applying the Minkowski inequality to \eqref{eq:sw_compute}\footnote{We apply PER only to the logit with a purpose of reducing $\parallel f^{\mW} \parallel$, while the original paper applies PER to every activation in a neural network.}. PER can be thought of as computing a robust norm in a randomly projected space induced by $\theta \sim P_\lambda$, which combines advantages of both the $L^1$ norm and the $L^2$ norm as well as captures the dependency between logits of each location by a random projection operation.


\section{Experiments} \label{experiment}
In this section, we examine the effectiveness of the function complexity regularization on improving the calibration performance. Throughout this section, we report mean values (rounded to two decimal places) obtained from five repetitions, and the experiments were performed on a single workstation with 8 GPUs (NVIDIA GeForce RTX 2080 Ti). We also note that all regularization methods do not result in a notable running time difference compared to the vanilla method, so we omit the running time comparison. To support reproducibility, we submit the code and will make it publicly available after the review period.

\subsection{Image Classification with ResNet and VGG}
We first consider the image-classification task on CIFAR \citep{krizhevsky2009learning}. We used the (pre-activation) ResNet \citep{he2016identity}, which is one of the most prevalent basis architectures in many state-of-the-art architectures \citep{huang2017densely,xie2017aggregated}. We also examined the VGG \citep{simonyan2015very} as a representative of models without a residual connection.

\textbf{Setup. }
We follow the training strategy presented in \citet{he2016identity}: we trained the model for 200 epochs by SGD with momentum coefficient 0.9, minibatch size of 128, and a weight decay rate 0.0001; weights were initialized as in \citet{he2015delving}; an initial learning rate was 0.1, and decreased by a factor of 10 at 100 and 150 epochs. We also used the augmentation presented in \citet{he2015delving}. In addition to the standard configuration, we used the initial learning rate warm-up, clipped the gradient when its norm exceeds one and made an extra validation set of 10,000 samples split from the training set. For convenience, we re-used this configuration for VGG-training except increasing the weight decay rate to 0.0005 as in \citet{simonyan2015very}.

We chose a regularization coefficient based on the validation set accuracy among the following four coefficients for each method: $\{ 0.1, 0.03, 0.01, 0.003 \}$ for $L^1$ norm and sliced Wasserstein regularizations; $\{ 0.03, 0.01, 0.003, 0.001 \}$ for $L^2$ norm regularization; $\{ 1.0, 0.3, 0.1, 0.03 \}$ for PER (10x lower for CIFAR-10).

We evaluate methods by classification error, negative log-likelihood (NLL), and ECE, which are commonly used in literature \citep{lakshminarayanan2017simple, guo2017calibration, snoek2019can}. We note that we do not use ECD as a performance measure since its role was to derive motivation for improving generalization of the training objective with explicit regularization and it has a strong correlation with ECE (cf. Figure~\ref{init_wd_test} (c)). 

\begin{table*}
\caption{Experimental results under various regularization methods on BERT. Arrows on the metrics represent the desirable direction. $\lambda^*$ represents the best hyperparameter. 
}
\label{benchmark_result_bert}
\begin{center}
\begin{small}
\begin{tabular}{llllll|lllll}
\toprule
& \multicolumn{5}{c |}{20-Newsgroup} & \multicolumn{5}{c}{Web of science} \\
method & Acc $\uparrow$ & NLL $\downarrow$ & ECE $\downarrow$ & $\parallel f^{\mW} \parallel_2$  & $\lambda^*$ & Acc $\uparrow$ & NLL $\downarrow$ & ECE $\downarrow$ & $\parallel f^{\mW} \parallel_2$ & $\lambda^*$
\\
\midrule
BERT & 84.51 & 0.70 & 6.40 & 9.71 & - &    81.28 & 0.94 & 8.26 & 20.13   &  -  \\
+ $\parallel f^{\mW} \parallel_1$ & 84.71 & 0.61 & 4.67 & 6.02  &  0.1 & 81.68 & 0.90 & 7.02 & 8.89  &  0.1    \\
+ $\parallel f^{\mW} \parallel_2^2$  & 85.04 & 0.64 & 3.24 & 5.11 & 0.1 & 81.83 & 0.94 & 5.79 & 7.70  &  0.1      \\
+ $SW_1(\mu^{\mW}_{\mathcal{D}^\prime}, \nu)$ & 84.76 & 0.67 & 3.78 & 4.98  &  0.1 & 81.44 & 0.89 & 7.22 & 11.33   &  0.1   \\
+ PER & 85.02 & 0.65 & 4.50 & 6.21 & 0.3   &  81.36 & 0.87 & 8.19 & 13.21  &  0.3    \\
\bottomrule
\end{tabular}
\end{small}
\end{center}
\vskip -0.1in
\end{table*}

\textbf{Results. }
Table~\ref{benchmark_result} lists the experimental results, in which both regularization in the function space and the Wasserstein probability space successfully reduced the function complexity, e.g., reduced the $L^2$ norm of ResNet by at least 34\% on CIFAR-10 and 68\% on CIFAR-100. We note that regularization methods reduce the function complexity without compromising the generalization performance, unlike weight decay; that is, all regularization methods achieves consistent improvements of test error rates. We also note that the sum of the Frobenius norm of weights often increases compared to the vanilla method and changes only at most 2\% when it decreases, which again shows the undesirability of adjusting the weight decay rate for the function complexity regularization.

More importantly, the predictive probability's quality under explicit regularization is significantly better than the quality under the vanilla method. For instance, the regularization methods reduce the NLL of ResNet by at least 13\% CIFAR-10 and 6\% on CIFAR-100 and reduce the ECE of ResNet by at least 19\% on CIFAR-10 and 41\% on CIFAR-100. These improvements are comparable to or better than those of temperature scaling. For instance, temperature scaling gives NLL of 1.15 and ECE of 8.41 on CIFAR-100\footnote{We split the test set into two equal-size sets (a performance measurement set and a temperature calibration set) and measure the performance of temperature scaling calibrated with the calibration set, and repeat the same procedure by reversing their roles. }.



We can also see that all regularization losses improve NLL, ECE, and accuracy of VGG, except $L^1$ regularization on CIFAR-100 (Table~\ref{benchmark_result}). However, the effectiveness of explicit regularization on VGG tends to be less significant compared to their effects on ResNet. This might be contributed to a small capacity of VGG, which can be inferred from that $\parallel f^{\mW} \parallel_2$ of VGG that is reduced by almost 50\% compared to ResNet.

\subsection{Document Classification with BERT}
We also perform experiments on the natural language processing domain in order to more thoroughly evaluate the effectiveness of explicit regularization. Inspired by the recent finding \citep{kong2020calibrated}, we perform document classification on 20 newsgroup dataset \citep{socher2012deep} and web of science dataset \citep{kowsari2017hdltex}. We obtain a classifier by adding a linear layer and a softmax to BERT \citep{devlin2018bert}. 

\textbf{Setup. }
Our configuration is based on \citet{kong2020calibrated}: we initialize the model by the pretrained BERT, and then the model is trained for 5 epochs by Adam optimizer \citep{kingma2014adam} with learning rate 0.00005 ($\beta_1=0.9, \beta_2=0.999$) and minibatch size of 32. 


The search spaces of regularization coefficients were:  $\{0.3, 0.1, 0.003, 0.001 \}$ for $L^1$ norm, $L^2$ norm sliced Wasserstein regularizations; $\{1.0, 0.3, 0.1, 0.003 \}$ for PER.

\textbf{Result. }
As consistent with the findings in the image classification task, all explicit regularization methods improve the reliability of predictive probability with consistent gains in test accuracy (Table \ref{benchmark_result_bert}). We believe the consistent results on two different domains (image classification and document classification) corroborate the significance of our findings, leaving extensive evaluation on more diverse environments as future work.


\begin{figure*}
    \centering
    \begin{subfigure}
     {\includegraphics[width=0.240\textwidth]{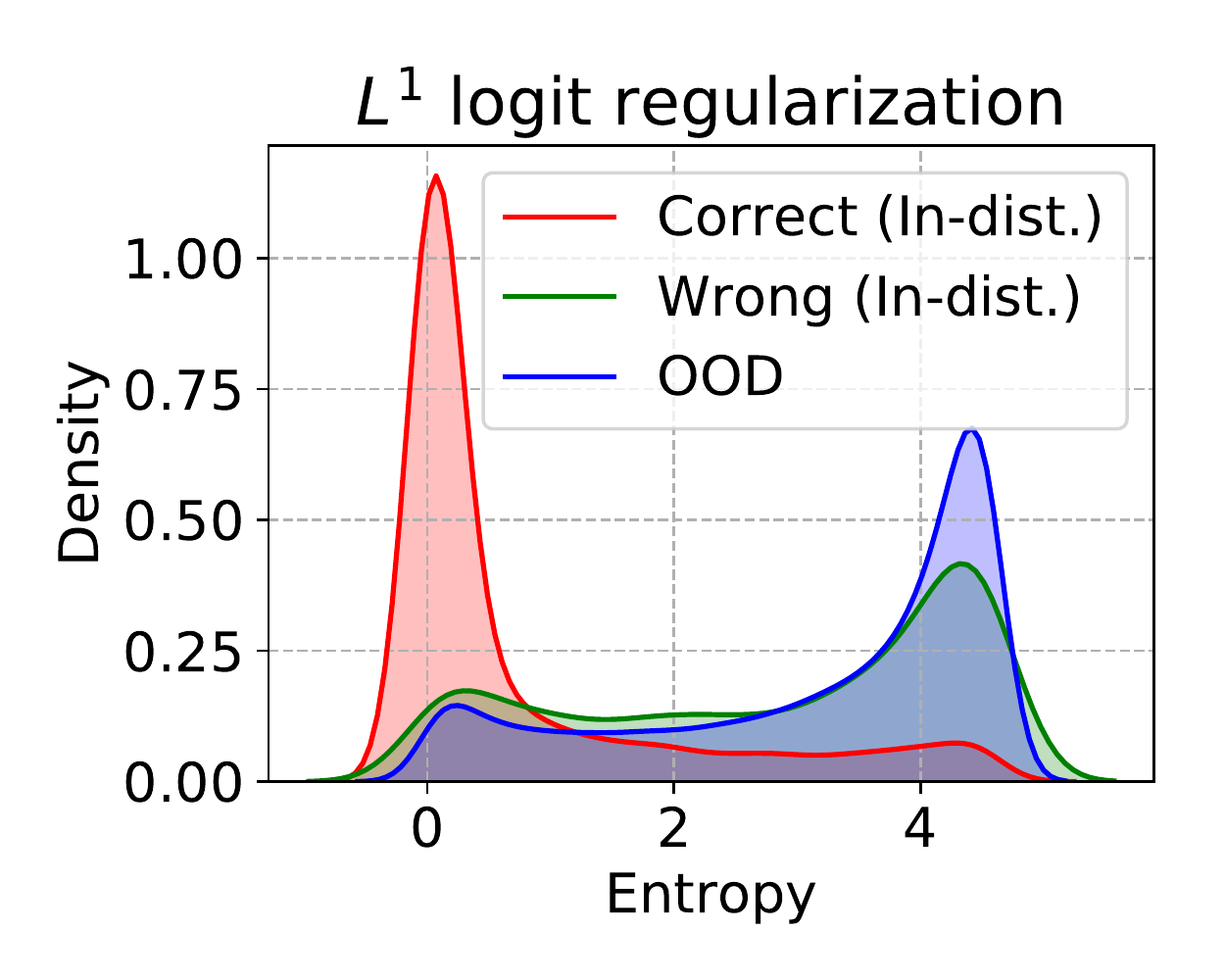}}
    \end{subfigure}
    \begin{subfigure}
     {\includegraphics[width=0.240\textwidth]{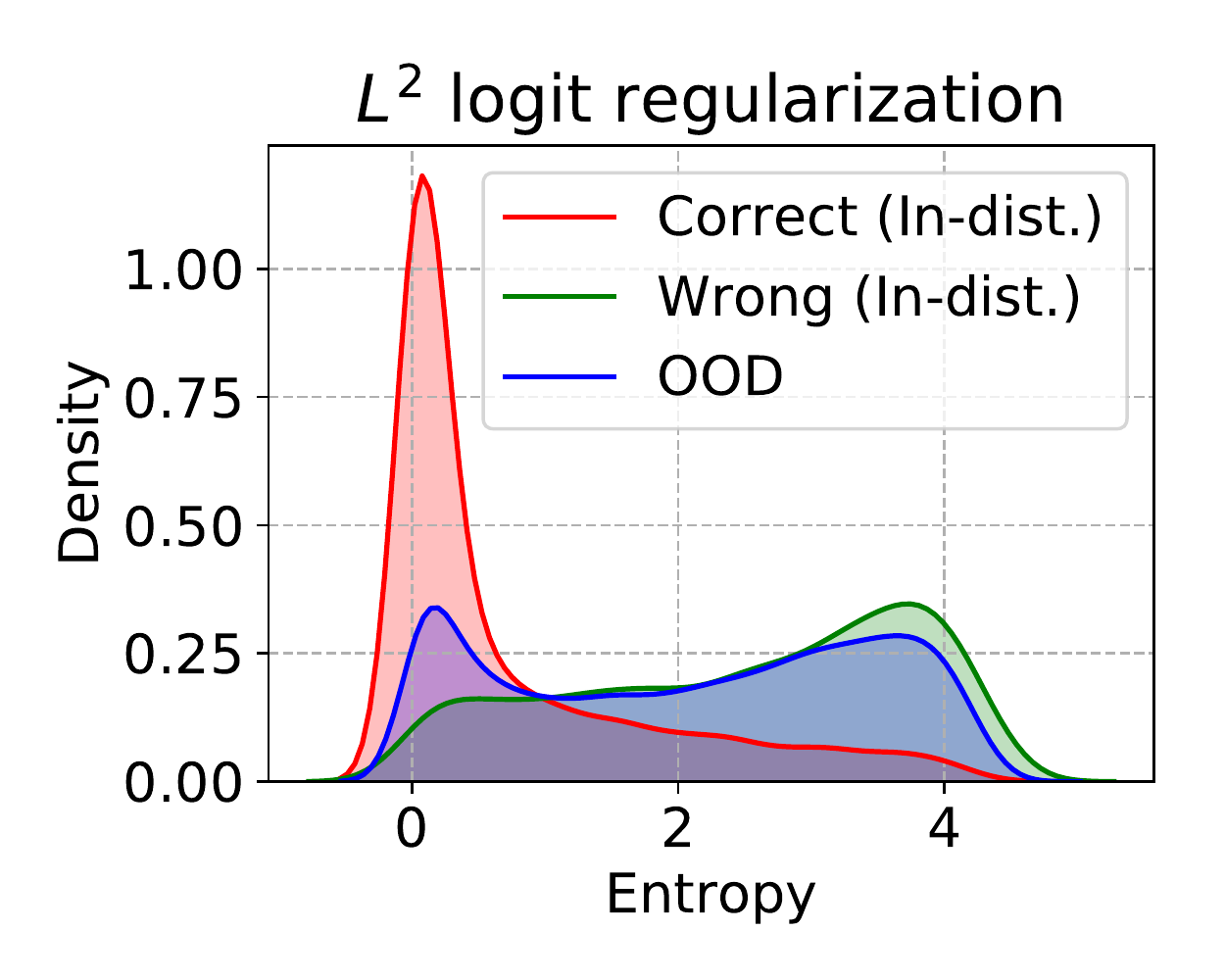}}
    \end{subfigure}
    \begin{subfigure}
     {\includegraphics[width=0.240\textwidth]{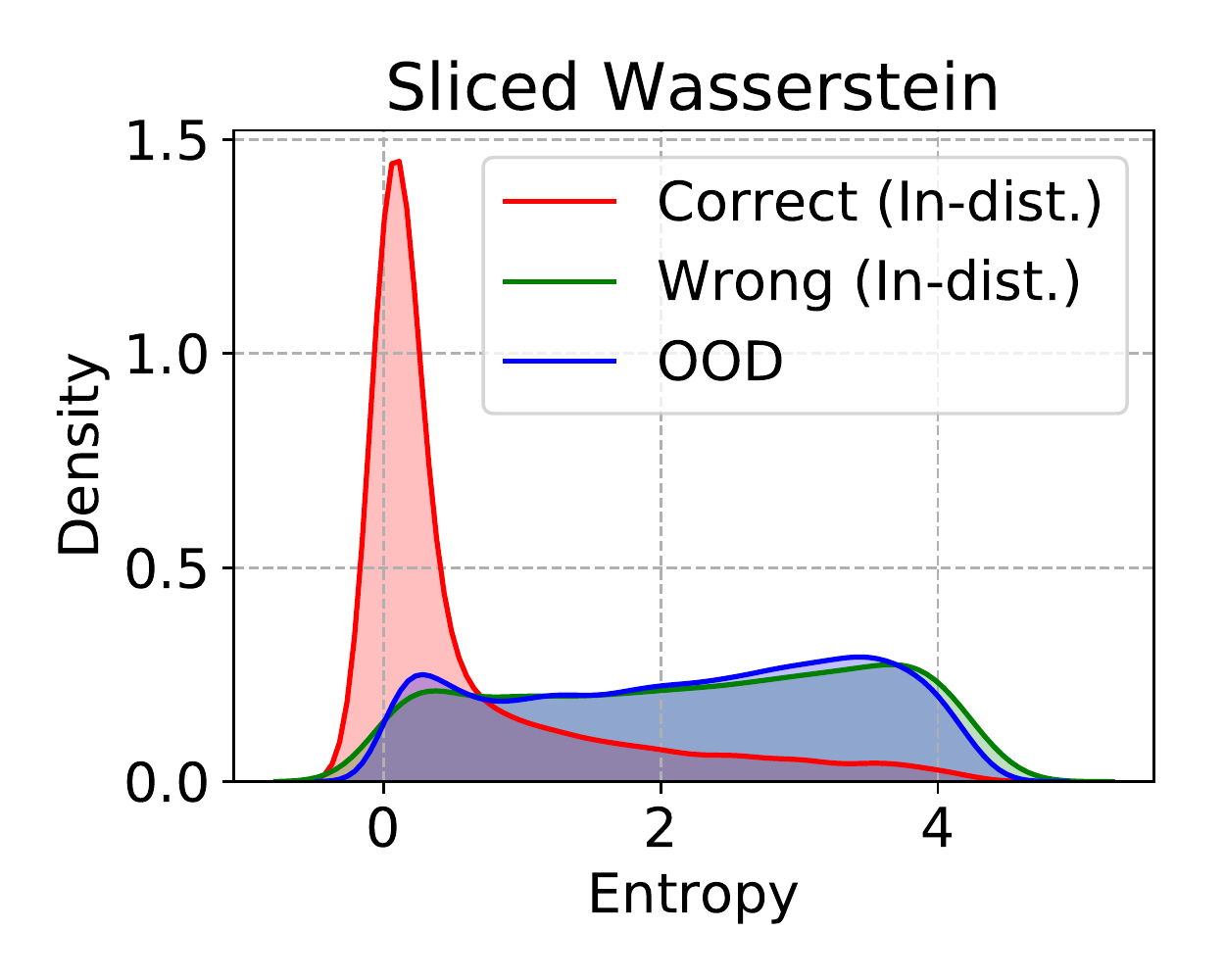}}
    \end{subfigure}
    \begin{subfigure}
     {\includegraphics[width=0.240\textwidth]{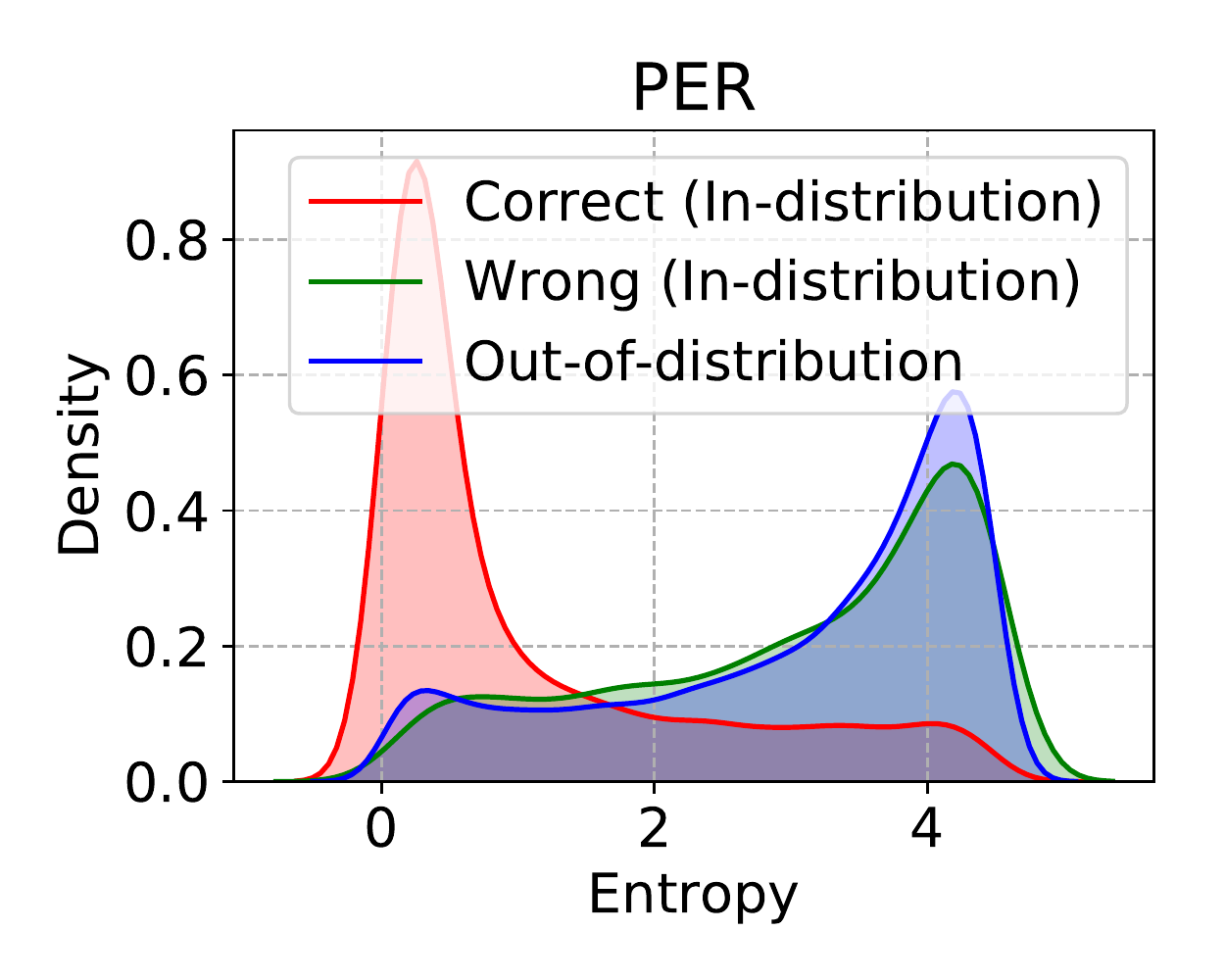}}
    \end{subfigure}
    \\
    \vskip -0.2in
    \begin{subfigure}
     {\includegraphics[width=0.240\textwidth]{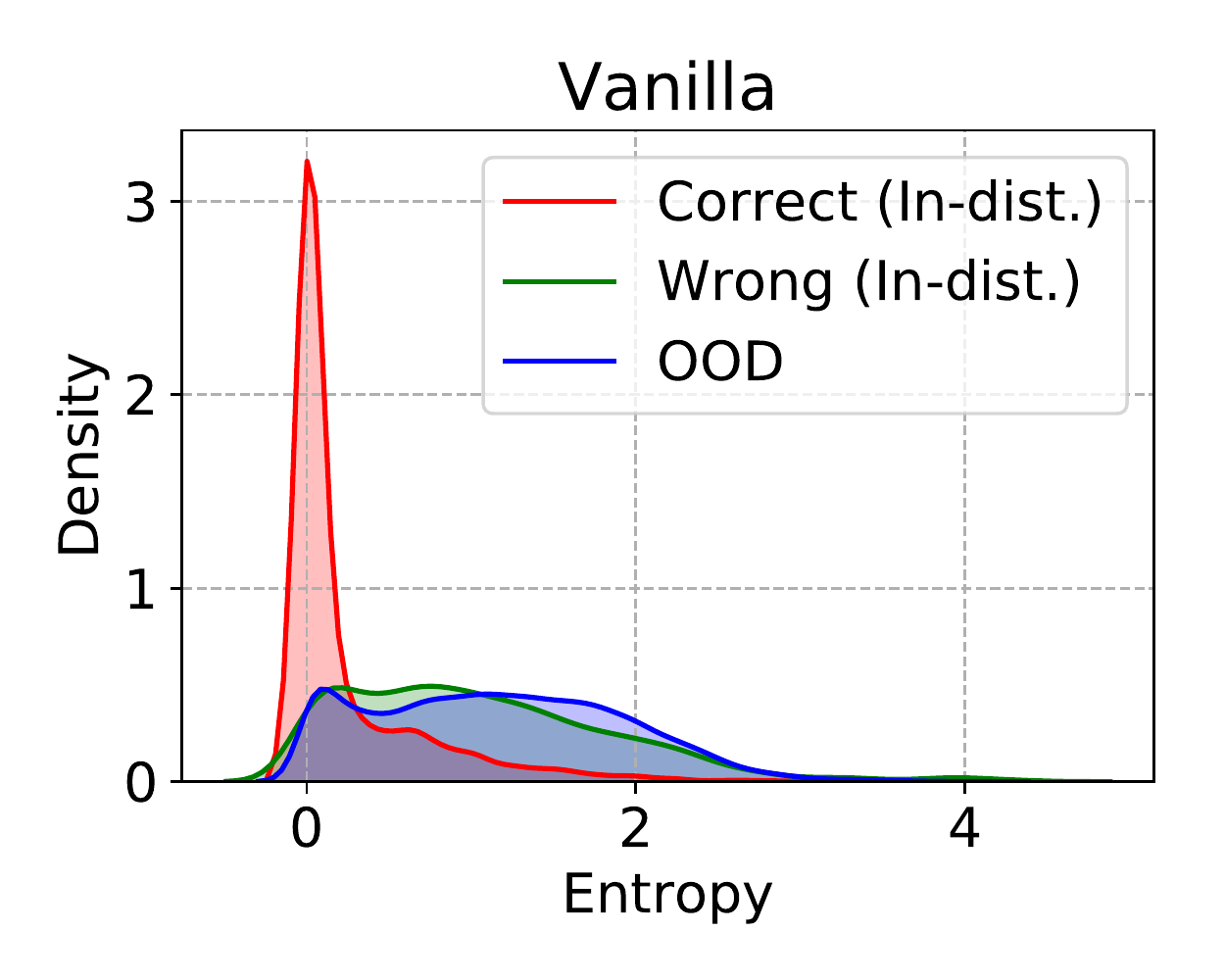}}
    \end{subfigure}
    \begin{subfigure}
     {\includegraphics[width=0.240\textwidth]{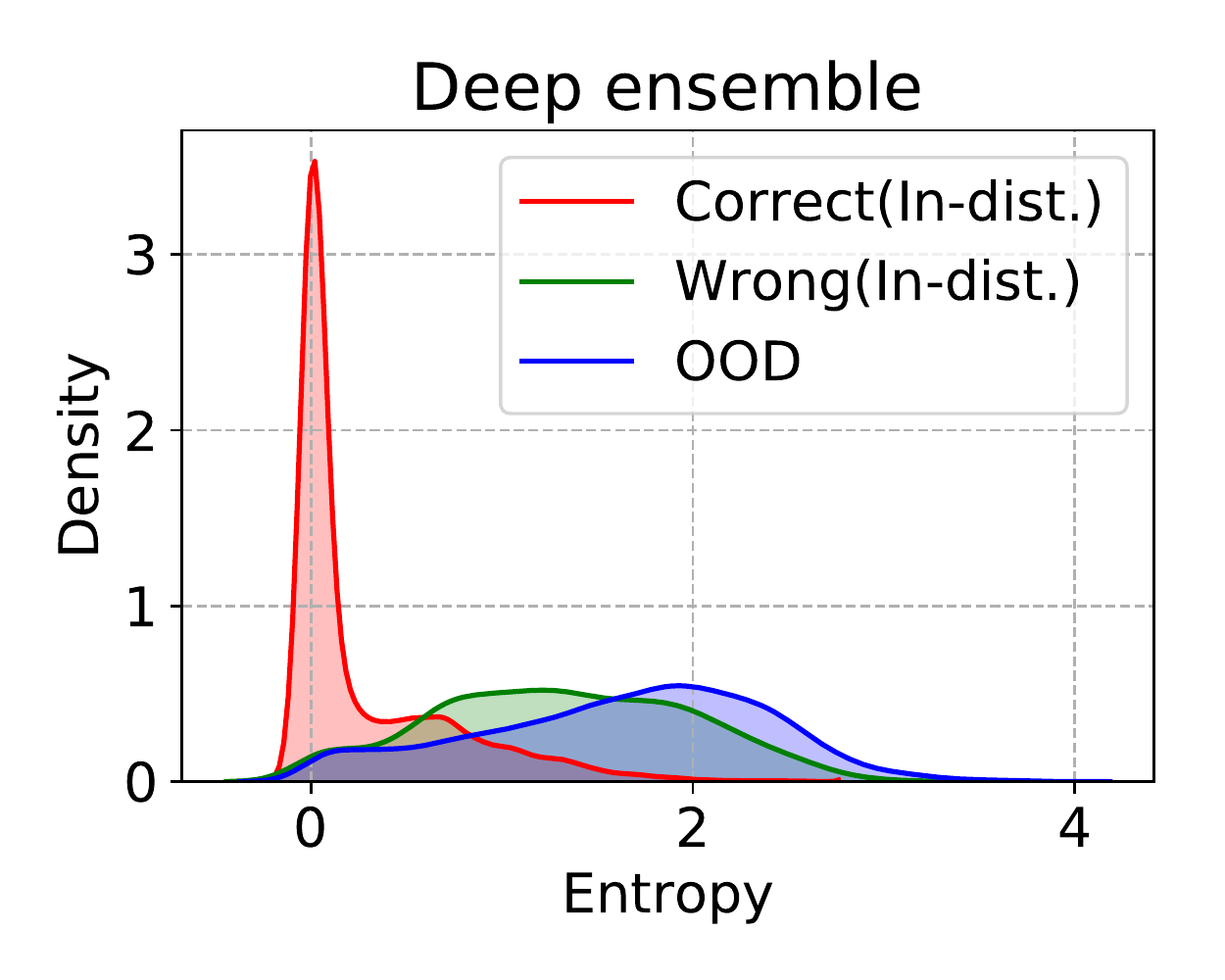}}
    \end{subfigure}
    \begin{subfigure}
     {\includegraphics[width=0.240\textwidth]{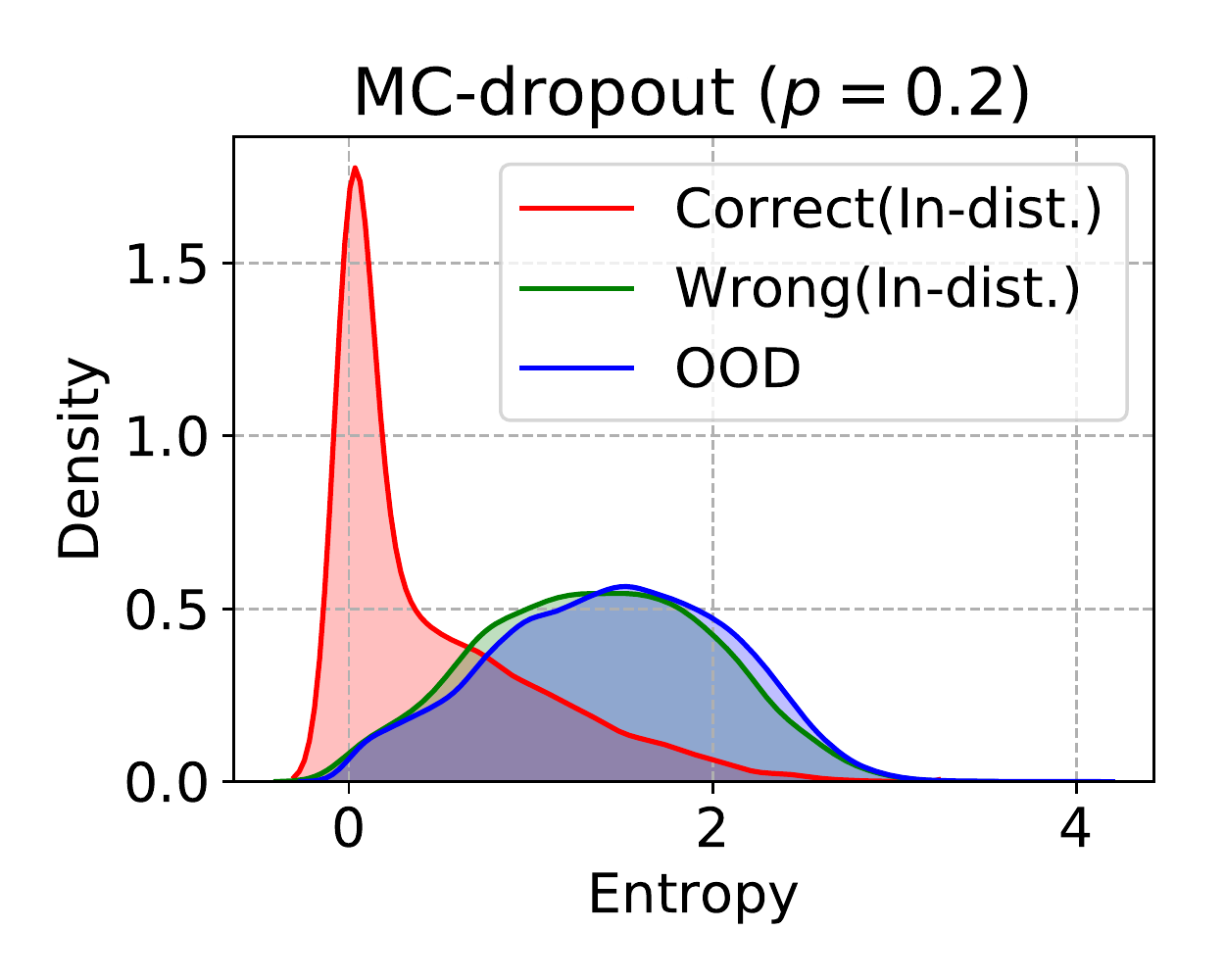}}
    \end{subfigure}
    \begin{subfigure}
     {\includegraphics[width=0.240\textwidth]{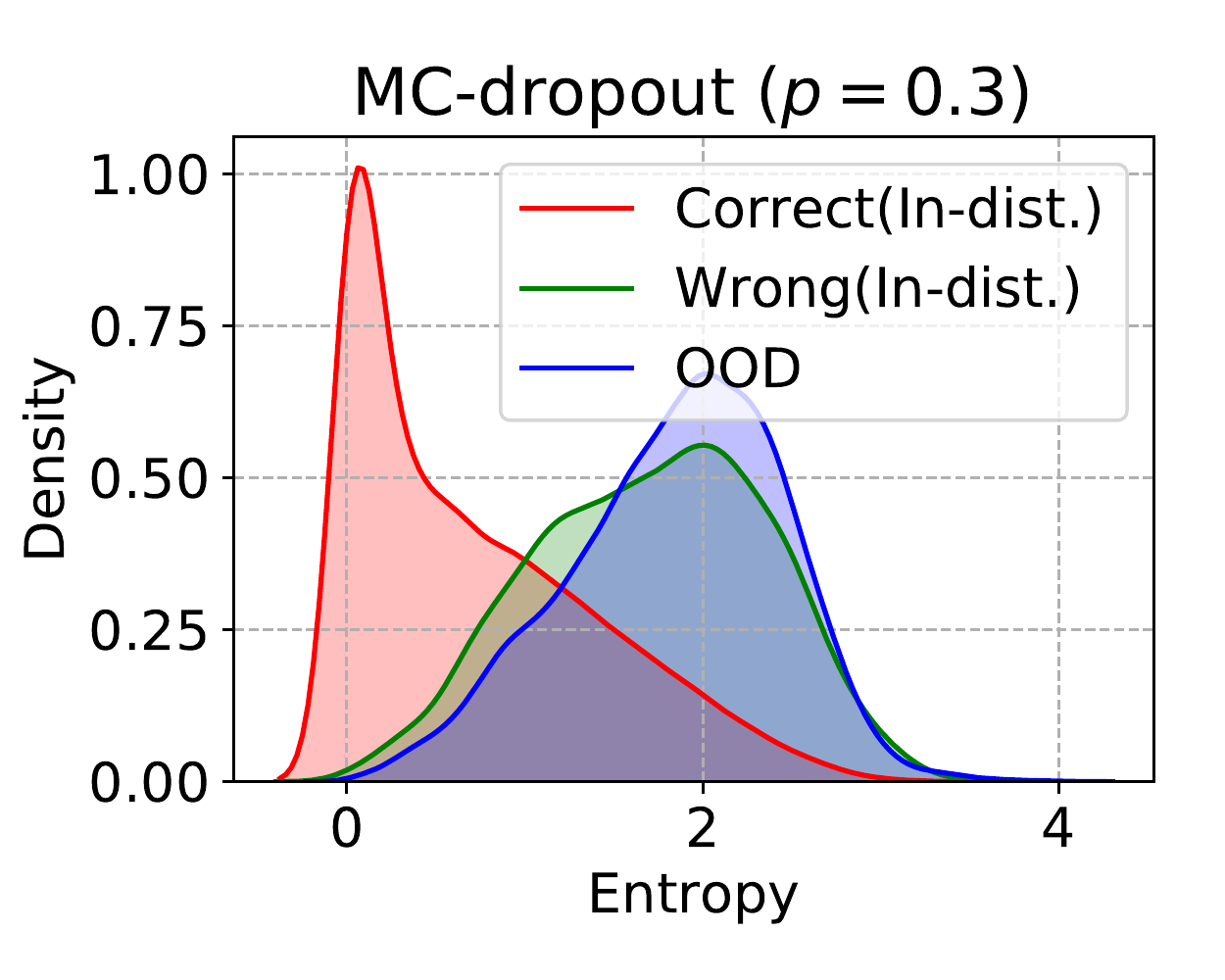}}
    \end{subfigure}
    \vskip -0.1in
    \caption{Density of predictive uncertainty on CIFAR-100 (in-distribution) and SVHN (OOD). The upper figures illustrate explicit regularization methods, and the lower figures 
    illustrate the vanilla method, ensemble methods, and Bayesian neural networks.}
    \vskip -0.1in
    \label{ood_plot}
\end{figure*}

\begin{table*}
\caption{Misclassification detection and OOD detection task performances based on NBAUCC$_{0.5}$. Arrows on the metrics represent the desirable direction. 
We present the result of MC-dropout with probability $0.3$, which performed best among $p \in \{0.2, 0.3, 0.5 \}$.
}
\label{quan_uncert}
\begin{center}
\begin{small}
\begin{tabular}{llllllll}
\toprule
Task & Vanilla & MC-dropout & Deep Ensemble &
$\parallel f^{\mW} \parallel_1$ & $\parallel f^{\mW} \parallel_2^2$ & $SW_1(\mu^{\mW}_{\mathcal{D}^\prime}, \nu)$ &    PER  \\
\midrule
Misclasification detection $\uparrow$ &  1.55 & 5.94 &  0.59 & 9.53 & 6.85 & 5.10 & 10.24     \\
OOD detection $\uparrow$ &  15.98 & 31.28 &  28.46 & 55.51 & 31.47 & 31.63 & 49.77 \\
\bottomrule
\end{tabular}
\end{small}
\end{center}
\vskip -0.1in
\end{table*}





\subsection{Predictive Uncertainty on Ignorant Samples}
We analyze the predictive uncertainty of ResNet-50 on misclassified samples and OOD samples, expecting it to produce the answer of ``I don't know'' for these samples (Figure~\ref{ood_plot}). We can see that the vanilla method provides somewhat low predictive entropy for OOD and misclassified samples, albeit the entropy is slightly higher than the entropy on correctly classified samples. In contrast, explicit regularization successfully gathers a mass of predictive entropy for both OOD samples and misclassified samples around the maximum-entropy region.

We compare predictive uncertainty under explicit regularization to those of Bayesian neural networks and ensemble methods. Specifically, we use the scalable Bayesian neural network, called MC-dropout \citep{gal2016dropout}, because other methods based on variational inference \citep{graves2011practical,blundell2015weight,wu2019deterministic} or MCMC \citep{welling2011bayesian,zhang2020cyclical} require modifications to the baseline including the optimization procedure and the architecture, which deters a fair comparison.  We searched a dropout rate over \{0.1, 0.2, 0.3, 0.4, 0.5 \} and used 100 number of Monte-Carlo samples at test time, i.e., 100x more inference time.  We also use the deep ensemble \citep{lakshminarayanan2017simple} with 5 number of ensembles, i.e., 5x more training and inference time. 

Figure~\ref{ood_plot} shows that the regularization-based methods produce significantly better uncertainty representation than the MC-dropout and deep ensemble; even though both deep ensemble and MC-dropout can move mass on less certain regions, the positions are still far from the highest uncertainty region, unlike the regularization-based methods.




For quantitative evaluation of the uncertainty representation ability, we measure misclassification and OOD detection performances. To this end, we use recently proposed normalized bounded area under the calibration curve (NBAUCC) \citep{kong2020calibrated}, which overcomes a disadvantage of classical measures such as AUROC and AUPRC that cannot consider the calibration performance. Specifically, NBAUCC with the upper bound of confidence threshold for misclassified or OOD samples, denoted by $\tau$, is computed by:
\begin{equation}
    \text{NBAUCC}_{\tau} = \frac{1}{M} \sum_{i=1}^{M} F_1\left(\frac{\tau}{M} i\right)
\end{equation}
where $F_1(t)$ computes $F_1$ score by regarding samples with predictive confidence higher than $\tau$ as correct (resp. in-distribution) samples and incorrect (resp. OOD) samples otherwise; $M$ is a predetermined step size.
We present the misclassification and OOD detection performances in Table~\ref{quan_uncert}, where all regularization methods significantly improve detection performances based on their better predictive uncertainty representation abilities.





\subsection{Comparison to Implicit Regularization}
Various forms of implicit regularization in deep learning have shown promising theoretical and empirical results on improving generalization performance \citep{hardt2015train, hoffer2017train,gunasekar2018implicit,li2019towards}. In this regard, we investigate their potential impacts on the calibration performance by magnifying implicit regularization effects of early stopping and SGD in ResNet-50.

\begin{figure}[]
  \centering
    \begin{subfigure}
        {\includegraphics[width=0.49\textwidth]{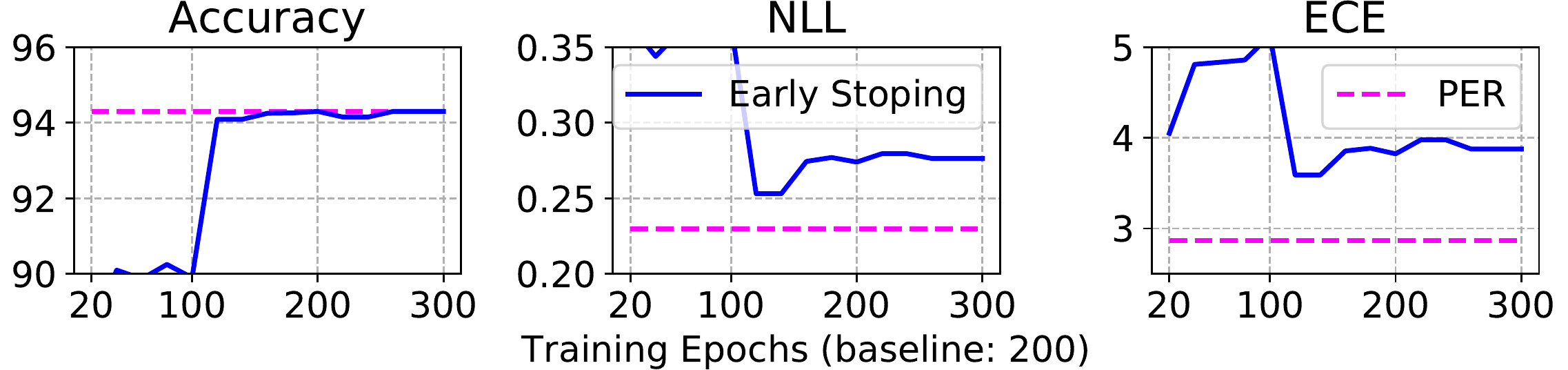}}
    \end{subfigure}
    \begin{subfigure}
        {\includegraphics[width=0.49\textwidth]{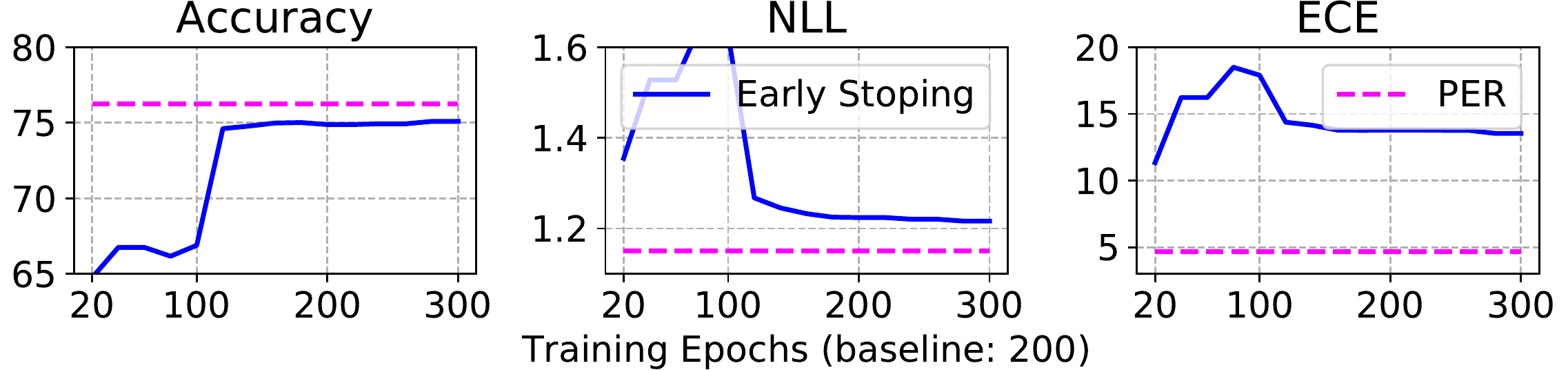}}
    \end{subfigure}
  \vskip -0.15in
  \caption{Impacts of early stopping on accuracy, negative log-likelihood, and ECE in CIFAR-10 (upper) and CIFAR-100 (lower). X-axis represents the maximum number of training epochs.}
  \label{implicit_reg}
\end{figure}

Figure~\ref{implicit_reg} shows that the early stopping does not result in dramatic improvements, unlike PER. On CIFAR-10, NLL and ECE are improved compared to the baseline, but these improvements are marginal compared to PER. On the contrary, on CIFAR-100, a longer training slightly benefits NLL and ECE due to increased classification accuracy. Considering that the $L^2$ norm achieves a significantly high value at the early stage of training (cf. Figure~\ref{train_analysis} (right)), this undesirable result may seem natural.

\begin{figure}
  \centering
    {\includegraphics[width=0.49\textwidth]{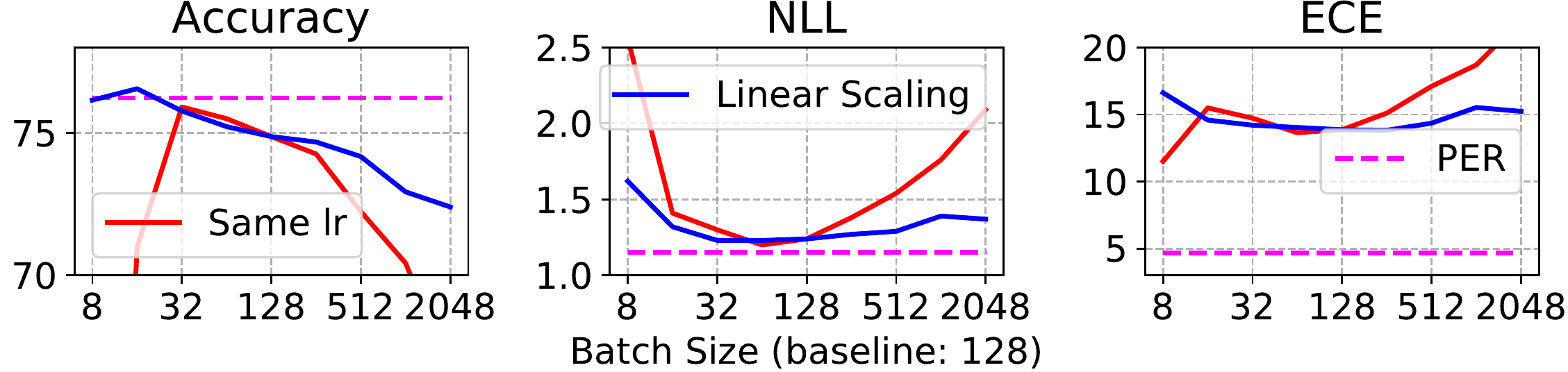}}
  \vskip -0.1in
  \caption{Impacts of varying noise in stochastic gradient computation on accuracy, negative log-likelihood, and ECE in CIFAR-100.}
  \label{implicit_reg_sgd}
\end{figure}

Next, we examine the effects of varying minibatch sizes under the baseline learning rate or linear scaling rule \citep{goyal2017accurate}, in which a smaller minibatch size increases the noise of the stochastic gradient. In Figure~\ref{implicit_reg_sgd}, we observe that increasing the noise level by using smaller batch sizes benefits generalization performance. However, it does not benefit and even worsens the predictive distribution quality (NLL and ECE). These observations highlight the importance of explicit regularization for improving the calibration performance of neural networks.  


\section{Related Work}
Recent works show that joint modeling of a generative model $p(\rvx)$ along with a classifier $p(\ry|\rvx)$, or $p(\rvx, \ry)$ directly, helps to produce calibrated predictive uncertainty \citep{alemi2018uncertainty,nalisnick2019hybrid,grathwohl2020your}. Specifically, \citet{alemi2018uncertainty} argue that modeling stochastic hidden representation through the variational information bottleneck principle \citep{alemi2016deep} allows representing better predictive uncertainty. This can be related to the effectiveness of ensemble methods for well-calibrated predictive uncertainty, which aggregate representations of several models \citep{lakshminarayanan2017simple,snoek2019can,ashukha2020pitfalls}. In this regard, hybrid modeling and ensemble methods share a similar principle to the Bayesian methods, concerning the \textit{stochasticity of the function}. However, this paper takes a fundamentally different approach that concentrates on explicit regularization in deterministic neural networks. 


Other works concentrate on \textit{the structural characteristics} of neural networks. Specifically, \citet{hein2019relu} identify the cause of the overconfidence problem based on an analysis of the affine compositional function, e.g., ReLU. The basic intuition behind this analysis is that one can always find a multiplier $\lambda$ to an input $\vx$, which makes a neural network produce one dominant entry on $\lambda \vx$. \citet{verma2019error} point out that the region of the highest predictive uncertainty under the softmax forms a subspace in the logit space, so the volume of an area representing high predictive uncertainty would be negligible. However, our approach suggests that these structural characteristics' inherent flaws can be easily cured by adding an explicit regularization term without changing the existing components of neural networks.

From the perspective of the statistical learning theory \citep{vapnik2013nature}, a regularization method minimizing some form of complexity measures, e.g., Rademacher complexity \citep{bartlett2005local} or VC-dimension \citep{vapnik2015uniform}, is required to achieve better generalization of overparameterized models by preventing memorization of intricate patterns existing only in training samples. However, the role of capacity control with explicit regularization is challenged by many observations in deep learning. Specifically, overparameterized neural networks achieve impressive generalization performance with only implicit regularizations contained in the optimization procedures \citep{hardt2015train,li2019towards} or the structural characteristics \citep{gunasekar2018implicit,hanin2019deep,luo2019towards}. Moreover, \citet{zhang2017understanding} show that explicit regularization does not prevent neural networks from fitting random labels that \textit{cannot be generalized} to unseen examples. Therefore, the importance of explicit regularization in deep learning seems to be questionable. In this work, we reemphasize its importance, presenting a different view on the role of regularization that improves predictive uncertainty representation.

\section{Conclusion}
In this works, we show the effectiveness of explicit regularization on improving the calibration of predictive uncertainty, which presents a novel view on the role and importance of explicit regularization. Specifically, our extensive experimental results show that explicit regularization methods improve calibration performance, predictive uncertainty representation, misclassification/OOD detection performances, and even test accuracy. The regularization-based approach has better computational efficiency and scalability compared to Bayesian neural networks and ensemble methods. Also, it is not subject to the quality of the holdout dataset, unlike post-hoc calibration methods. However, the regularization methods are limited in that they cannot utilize more sophisticated uncertainty quantification methods due to its deterministic nature, such as mutual information measuring epistemic uncertainty \citep{smith2018understanding}. We leave this limitation as an important future direction of research, which may be solved by using more expressive parameterization in the last layer, e.g., \citep{wilson2016kernel,skafte2019reliable,malinin2019reverse,joo2020being}.





\newpage

\bibliography{icml2021}

\begin{thebibliography}{59}
\providecommand{\natexlab}[1]{#1}
\providecommand{\url}[1]{\texttt{#1}}
\expandafter\ifx\csname urlstyle\endcsname\relax
  \providecommand{\doi}[1]{doi: #1}\else
  \providecommand{\doi}{doi: \begingroup \urlstyle{rm}\Url}\fi

\bibitem[Alemi et~al.(2017)Alemi, Fischer, Dillon, and Murphy]{alemi2016deep}
Alemi, A.~A., Fischer, I., Dillon, J.~V., and Murphy, K.
\newblock Deep variational information bottleneck.
\newblock In \emph{International Conference on Learning Representations}, 2017.

\bibitem[Alemi et~al.(2018)Alemi, Fischer, and Dillon]{alemi2018uncertainty}
Alemi, A.~A., Fischer, I., and Dillon, J.~V.
\newblock Uncertainty in the variational information bottleneck.
\newblock \emph{arXiv preprint arXiv:1807.00906}, 2018.

\bibitem[Ashukha et~al.(2020)Ashukha, Lyzhov, Molchanov, and
  Vetrov]{ashukha2020pitfalls}
Ashukha, A., Lyzhov, A., Molchanov, D., and Vetrov, D.
\newblock Pitfalls of in-domain uncertainty estimation and ensembling in deep
  learning.
\newblock In \emph{International Conference on Learning Representations}, 2020.

\bibitem[Bartlett et~al.(2005)Bartlett, Bousquet, Mendelson,
  et~al.]{bartlett2005local}
Bartlett, P.~L., Bousquet, O., Mendelson, S., et~al.
\newblock Local {R}ademacher complexities.
\newblock \emph{The Annals of Statistics}, 33\penalty0 (4):\penalty0
  1497--1537, 2005.

\bibitem[Blundell et~al.(2015)Blundell, Cornebise, Kavukcuoglu, and
  Wierstra]{blundell2015weight}
Blundell, C., Cornebise, J., Kavukcuoglu, K., and Wierstra, D.
\newblock Weight uncertainty in neural networks.
\newblock In \emph{International Conference on Machine Learning}, 2015.

\bibitem[Bridle(1990)]{bridle1990probabilistic}
Bridle, J.~S.
\newblock Probabilistic interpretation of feedforward classification network
  outputs, with relationships to statistical pattern recognition.
\newblock In \emph{Neurocomputing}, pp.\  227--236. Springer, 1990.

\bibitem[Dawid(1982)]{dawid1982well}
Dawid, A.~P.
\newblock The well-calibrated {B}ayesian.
\newblock \emph{Journal of the American Statistical Association}, 77\penalty0
  (379):\penalty0 605--610, 1982.

\bibitem[Devlin et~al.(2018)Devlin, Chang, Lee, and Toutanova]{devlin2018bert}
Devlin, J., Chang, M.-W., Lee, K., and Toutanova, K.
\newblock Bert: Pre-training of deep bidirectional transformers for language
  understanding.
\newblock \emph{arXiv preprint arXiv:1810.04805}, 2018.

\bibitem[Gal \& Ghahramani(2016)Gal and Ghahramani]{gal2016dropout}
Gal, Y. and Ghahramani, Z.
\newblock Dropout as a {B}ayesian approximation: Representing model uncertainty
  in deep learning.
\newblock In \emph{International Conference on Machine Learning}, 2016.

\bibitem[Goodfellow et~al.(2015)Goodfellow, Shlens, and
  Szegedy]{goodfellow2015explaining}
Goodfellow, I.~J., Shlens, J., and Szegedy, C.
\newblock Explaining and harnessing adversarial examples.
\newblock In \emph{International Conference on Learning Representations}, 2015.

\bibitem[Goyal et~al.(2017)Goyal, Doll{\'a}r, Girshick, Noordhuis, Wesolowski,
  Kyrola, Tulloch, Jia, and He]{goyal2017accurate}
Goyal, P., Doll{\'a}r, P., Girshick, R., Noordhuis, P., Wesolowski, L., Kyrola,
  A., Tulloch, A., Jia, Y., and He, K.
\newblock Accurate, large minibatch sgd: Training imagenet in 1 hour.
\newblock \emph{arXiv preprint arXiv:1706.02677}, 2017.

\bibitem[Grathwohl et~al.(2020)Grathwohl, Wang, Jacobsen, Duvenaud, Norouzi,
  and Swersky]{grathwohl2020your}
Grathwohl, W., Wang, K.-C., Jacobsen, J.-H., Duvenaud, D., Norouzi, M., and
  Swersky, K.
\newblock Your classifier is secretly an energy based model and you should
  treat it like one.
\newblock In \emph{International Conference on Learning Representations}, 2020.

\bibitem[Graves(2011)]{graves2011practical}
Graves, A.
\newblock Practical variational inference for neural networks.
\newblock In \emph{Advances in Neural Information Processing Systems}, 2011.

\bibitem[Gunasekar et~al.(2018)Gunasekar, Lee, Soudry, and
  Srebro]{gunasekar2018implicit}
Gunasekar, S., Lee, J.~D., Soudry, D., and Srebro, N.
\newblock Implicit bias of gradient descent on linear convolutional networks.
\newblock In \emph{Advances in Neural Information Processing Systems}, 2018.

\bibitem[Guo et~al.(2017)Guo, Pleiss, Sun, and Weinberger]{guo2017calibration}
Guo, C., Pleiss, G., Sun, Y., and Weinberger, K.~Q.
\newblock On calibration of modern neural networks.
\newblock In \emph{International Conference on Machine Learning}, 2017.

\bibitem[Hanin \& Rolnick(2019)Hanin and Rolnick]{hanin2019deep}
Hanin, B. and Rolnick, D.
\newblock Deep relu networks have surprisingly few activation patterns.
\newblock In \emph{Advances in Neural Information Processing Systems}, 2019.

\bibitem[Hardt et~al.(2016)Hardt, Recht, and Singer]{hardt2015train}
Hardt, M., Recht, B., and Singer, Y.
\newblock Train faster, generalize better: Stability of stochastic gradient
  descent.
\newblock In \emph{International Conference on Machine Learning}, 2016.

\bibitem[He et~al.(2015)He, Zhang, Ren, and Sun]{he2015delving}
He, K., Zhang, X., Ren, S., and Sun, J.
\newblock Delving deep into rectifiers: Surpassing human-level performance on
  imagenet classification.
\newblock In \emph{IEEE International Conference on Computer Vision}, 2015.

\bibitem[He et~al.(2016)He, Zhang, Ren, and Sun]{he2016identity}
He, K., Zhang, X., Ren, S., and Sun, J.
\newblock Identity mappings in deep residual networks.
\newblock In \emph{European Conference on Computer Vision}, pp.\  630--645,
  2016.

\bibitem[Hein et~al.(2019)Hein, Andriushchenko, and Bitterwolf]{hein2019relu}
Hein, M., Andriushchenko, M., and Bitterwolf, J.
\newblock Why relu networks yield high-confidence predictions far away from the
  training data and how to mitigate the problem.
\newblock In \emph{IEEE Conference on Computer Vision and Pattern Recognition},
  2019.

\bibitem[Hoffer et~al.(2017)Hoffer, Hubara, and Soudry]{hoffer2017train}
Hoffer, E., Hubara, I., and Soudry, D.
\newblock Train longer, generalize better: closing the generalization gap in
  large batch training of neural networks.
\newblock In \emph{Advances in Neural Information Processing Systems}, 2017.

\bibitem[Huang et~al.(2017)Huang, Liu, Van Der~Maaten, and
  Weinberger]{huang2017densely}
Huang, G., Liu, Z., Van Der~Maaten, L., and Weinberger, K.~Q.
\newblock Densely connected convolutional networks.
\newblock In \emph{IEEE Conference on Computer Vision and Pattern Recognition},
  2017.

\bibitem[Joo et~al.(2020{\natexlab{a}})Joo, Chung, and Seo]{joo2020being}
Joo, T., Chung, U., and Seo, M.-G.
\newblock Being {B}ayesian about categorical probability.
\newblock In \emph{International Conference on Machine Learning},
  2020{\natexlab{a}}.

\bibitem[Joo et~al.(2020{\natexlab{b}})Joo, Kang, and Kim]{joo2020regularizing}
Joo, T., Kang, D., and Kim, B.
\newblock Regularizing activations in neural networks via distribution matching
  with the {W}asserstein metric.
\newblock In \emph{International Conference on Learning Representations},
  2020{\natexlab{b}}.

\bibitem[Kingma \& Ba(2015)Kingma and Ba]{kingma2014adam}
Kingma, D.~P. and Ba, J.
\newblock Adam: A method for stochastic optimization.
\newblock In \emph{International Conference on Learning Representations}, 2015.

\bibitem[Kong et~al.(2020)Kong, Jiang, Zhuang, Lyu, Zhao, and
  Zhang]{kong2020calibrated}
Kong, L., Jiang, H., Zhuang, Y., Lyu, J., Zhao, T., and Zhang, C.
\newblock Calibrated language model fine-tuning for in-and out-of-distribution
  data.
\newblock In \emph{Conference on Empirical Methods in Natural Language
  Processing}, 2020.

\bibitem[Kowsari et~al.(2017)Kowsari, Brown, Heidarysafa, Meimandi, Gerber, and
  Barnes]{kowsari2017hdltex}
Kowsari, K., Brown, D.~E., Heidarysafa, M., Meimandi, K.~J., Gerber, M.~S., and
  Barnes, L.~E.
\newblock Hdltex: Hierarchical deep learning for text classification.
\newblock In \emph{IEEE International Conference on Machine Learning and
  Applications}, pp.\  364--371. IEEE, 2017.

\bibitem[Krizhevsky et~al.(2009)Krizhevsky, Hinton,
  et~al.]{krizhevsky2009learning}
Krizhevsky, A., Hinton, G., et~al.
\newblock Learning multiple layers of features from tiny images.
\newblock 2009.

\bibitem[Krogh \& Hertz(1992)Krogh and Hertz]{krogh1992simple}
Krogh, A. and Hertz, J.~A.
\newblock A simple weight decay can improve generalization.
\newblock In \emph{Advances in Neural Information Processing Systems}, 1992.

\bibitem[Lakshminarayanan et~al.(2017)Lakshminarayanan, Pritzel, and
  Blundell]{lakshminarayanan2017simple}
Lakshminarayanan, B., Pritzel, A., and Blundell, C.
\newblock Simple and scalable predictive uncertainty estimation using deep
  ensembles.
\newblock In \emph{Advances in Neural Information Processing Systems}, 2017.

\bibitem[Li et~al.(2019)Li, Wei, and Ma]{li2019towards}
Li, Y., Wei, C., and Ma, T.
\newblock Towards explaining the regularization effect of initial large
  learning rate in training neural networks.
\newblock In \emph{Advances in Neural Information Processing Systems}, 2019.

\bibitem[Loshchilov \& Hutter(2019)Loshchilov and
  Hutter]{loshchilov2017decoupled}
Loshchilov, I. and Hutter, F.
\newblock Decoupled weight decay regularization.
\newblock In \emph{International Conference on Learning Representations}, 2019.

\bibitem[Luo et~al.(2019)Luo, Wang, Shao, and Peng]{luo2019towards}
Luo, P., Wang, X., Shao, W., and Peng, Z.
\newblock Towards understanding regularization in batch normalization.
\newblock In \emph{International Conference on Learning Representations}, 2019.

\bibitem[MacKay(1992)]{mackay1992practical}
MacKay, D.~J.
\newblock A practical {B}ayesian framework for backpropagation networks.
\newblock \emph{Neural Computation}, 4\penalty0 (3):\penalty0 448--472, 1992.

\bibitem[Malinin \& Gales(2019)Malinin and Gales]{malinin2019reverse}
Malinin, A. and Gales, M.
\newblock Reverse {KL}-divergence training of prior networks: Improved
  uncertainty and adversarial robustness.
\newblock In \emph{Advances in Neural Information Processing Systems}, 2019.

\bibitem[M{\"u}ller et~al.(2019)M{\"u}ller, Kornblith, and
  Hinton]{muller2019does}
M{\"u}ller, R., Kornblith, S., and Hinton, G.
\newblock When does label smoothing help?
\newblock In \emph{Advances in Neural Information Processing Systems}, 2019.

\bibitem[Naeini et~al.(2015)Naeini, Cooper, and
  Hauskrecht]{naeini2015obtaining}
Naeini, M.~P., Cooper, G., and Hauskrecht, M.
\newblock Obtaining well calibrated probabilities using {B}ayesian binning.
\newblock In \emph{AAAI Conference on Artificial Intelligence}, 2015.

\bibitem[Nalisnick et~al.(2019)Nalisnick, Matsukawa, Teh, Gorur, and
  Lakshminarayanan]{nalisnick2019hybrid}
Nalisnick, E., Matsukawa, A., Teh, Y.~W., Gorur, D., and Lakshminarayanan, B.
\newblock Hybrid models with deep and invertible features.
\newblock In \emph{International Conference on Machine Learning}, 2019.

\bibitem[Neal(1993)]{neal1993bayesian}
Neal, R.~M.
\newblock Bayesian learning via stochastic dynamics.
\newblock In \emph{Advances in Neural Information Processing Systems}, 1993.

\bibitem[Osawa et~al.(2019)Osawa, Swaroop, Khan, Jain, Eschenhagen, Turner, and
  Yokota]{osawa2019practical}
Osawa, K., Swaroop, S., Khan, M. E.~E., Jain, A., Eschenhagen, R., Turner,
  R.~E., and Yokota, R.
\newblock Practical deep learning with {B}ayesian principles.
\newblock In \emph{Advances in Neural Information Processing Systems}, 2019.

\bibitem[Ovadia et~al.(2019)Ovadia, Fertig, Ren, Nado, Sculley, Nowozin,
  Dillon, Lakshminarayanan, and Snoek]{snoek2019can}
Ovadia, Y., Fertig, E., Ren, J., Nado, Z., Sculley, D., Nowozin, S., Dillon,
  J., Lakshminarayanan, B., and Snoek, J.
\newblock Can you trust your model's uncertainty? evaluating predictive
  uncertainty under dataset shift.
\newblock In \emph{Advances in Neural Information Processing Systems}, 2019.

\bibitem[Peyr{\'e} et~al.(2019)Peyr{\'e}, Cuturi,
  et~al.]{peyre2019computational}
Peyr{\'e}, G., Cuturi, M., et~al.
\newblock Computational optimal transport.
\newblock \emph{Foundations and Trends{\textregistered} in Machine Learning},
  11\penalty0 (5-6):\penalty0 355--607, 2019.

\bibitem[Robbins \& Monro(1951)Robbins and Monro]{robbins1951stochastic}
Robbins, H. and Monro, S.
\newblock A stochastic approximation method.
\newblock \emph{The Annals of Mathematical Statistics}, pp.\  400--407, 1951.

\bibitem[Simonyan \& Zisserman(2015)Simonyan and Zisserman]{simonyan2015very}
Simonyan, K. and Zisserman, A.
\newblock Very deep convolutional networks for large-scale image recognition.
\newblock In \emph{International Conference on Learning Representations}, 2015.

\bibitem[Skafte et~al.(2019)Skafte, J{\o}rgensen, and
  Hauberg]{skafte2019reliable}
Skafte, N., J{\o}rgensen, M., and Hauberg, S.
\newblock Reliable training and estimation of variance networks.
\newblock In \emph{Advances in Neural Information Processing Systems}, 2019.

\bibitem[Smith \& Gal(2018)Smith and Gal]{smith2018understanding}
Smith, L. and Gal, Y.
\newblock Understanding measures of uncertainty for adversarial example
  detection.
\newblock \emph{arXiv preprint arXiv:1803.08533}, 2018.

\bibitem[Socher et~al.(2012)Socher, Bengio, and Manning]{socher2012deep}
Socher, R., Bengio, Y., and Manning, C.~D.
\newblock Deep learning for nlp (without magic).
\newblock In \emph{Tutorial Abstracts of ACL 2012}, pp.\  5--5. 2012.

\bibitem[Szegedy et~al.(2016)Szegedy, Vanhoucke, Ioffe, Shlens, and
  Wojna]{szegedy2016rethinking}
Szegedy, C., Vanhoucke, V., Ioffe, S., Shlens, J., and Wojna, Z.
\newblock Rethinking the inception architecture for computer vision.
\newblock In \emph{IEEE Conference on Computer Vision and Pattern Recognition},
  2016.

\bibitem[Thulasidasan et~al.(2019)Thulasidasan, Chennupati, Bilmes,
  Bhattacharya, and Michalak]{thulasidasan2019mixup}
Thulasidasan, S., Chennupati, G., Bilmes, J.~A., Bhattacharya, T., and
  Michalak, S.
\newblock On mixup training: Improved calibration and predictive uncertainty
  for deep neural networks.
\newblock In \emph{Advances in Neural Information Processing Systems}, 2019.

\bibitem[Vapnik(1995)]{vapnik2013nature}
Vapnik, V.
\newblock \emph{The Nature of Statistical Learning Theory}.
\newblock Springer-Verlag, 1995.

\bibitem[Vapnik \& Chervonenkis(2015)Vapnik and
  Chervonenkis]{vapnik2015uniform}
Vapnik, V.~N. and Chervonenkis, A.~Y.
\newblock On the uniform convergence of relative frequencies of events to their
  probabilities.
\newblock In \emph{Measures of Complexity}, pp.\  11--30. Springer, 2015.

\bibitem[Verma \& Swami(2019)Verma and Swami]{verma2019error}
Verma, G. and Swami, A.
\newblock Error correcting output codes improve probability estimation and
  adversarial robustness of deep neural networks.
\newblock In \emph{Advances in Neural Information Processing Systems}, 2019.

\bibitem[Welling \& Teh(2011)Welling and Teh]{welling2011bayesian}
Welling, M. and Teh, Y.~W.
\newblock Bayesian learning via stochastic gradient {L}angevin dynamics.
\newblock In \emph{International Conference on Machine Learning}, 2011.

\bibitem[Wilson et~al.(2016)Wilson, Hu, Salakhutdinov, and
  Xing]{wilson2016kernel}
Wilson, A.~G., Hu, Z., Salakhutdinov, R.~R., and Xing, E.~P.
\newblock Deep kernel learning.
\newblock In \emph{International Conference on Artificial Intelligence and
  Statistics}, 2016.

\bibitem[Wu et~al.(2019)Wu, Nowozin, Meeds, Turner, Hern{\'a}ndez-Lobato, and
  Gaunt]{wu2019deterministic}
Wu, A., Nowozin, S., Meeds, E., Turner, R.~E., Hern{\'a}ndez-Lobato, J.~M., and
  Gaunt, A.~L.
\newblock Deterministic variational inference for robust {B}ayesian neural
  networks.
\newblock In \emph{International Conference on Learning Representations}, 2019.

\bibitem[Xie et~al.(2017)Xie, Girshick, Doll{\'a}r, Tu, and
  He]{xie2017aggregated}
Xie, S., Girshick, R., Doll{\'a}r, P., Tu, Z., and He, K.
\newblock Aggregated residual transformations for deep neural networks.
\newblock In \emph{IEEE Conference on Computer Vision and Pattern Recognition},
  2017.

\bibitem[Zhang et~al.(2017)Zhang, Bengio, Hardt, Recht, and
  Vinyals]{zhang2017understanding}
Zhang, C., Bengio, S., Hardt, M., Recht, B., and Vinyals, O.
\newblock Understanding deep learning requires rethinking generalization.
\newblock In \emph{International Conference on Learning Representations}, 2017.

\bibitem[Zhang et~al.(2018)Zhang, Cisse, Dauphin, and
  Lopez-Paz]{zhang2017mixup}
Zhang, H., Cisse, M., Dauphin, Y.~N., and Lopez-Paz, D.
\newblock Mixup: Beyond empirical risk minimization.
\newblock In \emph{International Conference on Learning Representations}, 2018.

\bibitem[Zhang et~al.(2020)Zhang, Li, Zhang, Chen, and
  Wilson]{zhang2020cyclical}
Zhang, R., Li, C., Zhang, J., Chen, C., and Wilson, A.~G.
\newblock Cyclical stochastic gradient {MCMC} for {B}ayesian deep learning.
\newblock In \emph{International Conference on Learning Representations}, 2020.

\end{thebibliography}
\bibliographystyle{icml2021}

\end{document}